\documentclass[letterpaper,10pt]{IEEEconf} 

\usepackage{graphicx}
\usepackage{dcolumn}
\usepackage{bm}
\usepackage{newtxtext, newtxmath} 

\usepackage{listings}
\usepackage{url}
\usepackage{amsmath}
\usepackage{amssymb}
\usepackage{color}
\usepackage{array}
\usepackage[capitalise]{cleveref}
\usepackage{tabularx, booktabs}
\usepackage{subfigure}
\usepackage{csquotes}

\definecolor{pgreen}{rgb}{0,0.5,0}
\definecolor{comment}{rgb}{0.4,0.4,0.4}
\newcommand{\ie}{i.\,e.}
\newcommand{\eg}{e.\,g.}
\newcommand{\vadere}{Vadere}
\newcommand{\code}[1]{{\footnotesize\ttfamily #1}}

\DeclareMathOperator*{\argmin}{arg\,min}
\DeclareMathOperator*{\disc}{disc}

\title{\vadere{}: An open-source simulation framework to promote interdisciplinary understanding}

\author{%
	Benedikt Kleinmeier$^{1,2,a}$, Benedikt Z\"{o}nnchen$^{1,2,b}$, Marion G\"{o}del$^{1,2}$, Gerta K\"{o}ster$^{1}$\\
	\begin{affiliation}
		$^{1}$ Munich University of Applied Sciences, Department of Computer Science and Mathematics,\\80335 Munich, Germany
	\end{affiliation}\\
	\begin{affiliation}
		$^{2}$ Technical University of Munich, Department of Informatics,\\85748 Garching, Germany
	\end{affiliation}\\
E-Mail: $^{a}$ \email{benedikt.kleinmeier@hm.edu},  $^{b}$ \email{zoennchen.benedikt@hm.edu}
}

\begin{document}
\maketitle
\thispagestyle{empty}
\pagestyle{empty}
\begin{abstract}
Pedestrian dynamics is an interdisciplinary field of research. Psychologists, sociologists, traffic engineers, physicists, mathematicians and computer scientists all strive to understand the dynamics of a moving crowd. 
In principle, computer simulations offer means to further this understanding. Yet, unlike for many classic dynamical systems in physics, there is no universally accepted locomotion model for crowd dynamics. On the contrary, a multitude of approaches, with very different characteristics, compete. Often only the experts in one special model type are able to assess the consequences these characteristics have on a simulation study.
Therefore, scientists from all disciplines who wish to use 
simulations to analyze pedestrian dynamics need a tool to compare competing approaches. Developers, too, would profit from an easy way to get insight into an alternative modeling ansatz. \vadere{} meets this interdisciplinary demand by offering an open-source
simulation framework that is lightweight in its approach and in its user interface while offering pre-implemented versions of the most widely spread models.
\end{abstract}

\maketitle


\section{Introduction}
\label{sec:intro}

Pedestrian dynamics is an active and versatile research area that attracts scientists from sociology and psychology to engineering, computer science and mathematics. This interdisciplinary scientific community shares a common goal: 
to enhance the understanding of crowd behavior.
Scientists approach this goal from different angles:
For example, sociologists and psychologists observe and analyze the human behaviors that affect pedestrian movement \cite{drury-2010}. They describe their findings verbally. Mathematicians and physicists mold the observed behavior into equations \cite{helbing-1995}. Computer scientists develop computer models based on these equations \cite{seitz-2016}. Each researcher wants to get new insights by using different but connected techniques and by asking different but connected questions.

The interdisciplinary character of pedestrian dynamics entails communication problems and misunderstandings: For example, the term \enquote{agent} \cite{adrian-2019} used by computer scientists to denote a simulated pedestrian may not be immediately understood as such by psychologists. The term \enquote{crowd} has a very rich meaning for social psychologists \cite{reicher-1984,drury-1999,challenger-2009} while a physicist may simply think of an aggregation of individuals like an aggregation of particles \cite{templeton-2015}. A third prominent example is the term \enquote{panic}. While panic is considered as myth by sociologist and human behaviour scientists \cite{fahy-2012}, this word is still used by many computer scientists \cite{helbing-2000,helbing-2012}. Thus terms might even be misleading in an interdisciplinary group. The consequences of these language difficulties go beyond research as soon as decision makers at crowded events, such as  city planners and the police apply research results. 

A pedestrian simulation tool reflects the interdisciplinary character of the field. Behaviors that were observed and described by empirical scientists are operationalized, then mapped to equations, algorithms, and finally computer programs. These computer programs can be executed to reenact real scenarios in a virtual world, where they can be re-observed.
Therefore, they are a powerful tool to foster interdisciplinary understanding in pedestrian dynamics.

Laboratory and field experiments are essential to understand pedestrian behavior, but they need participants and bind a lot of scientific and non-scientific human resources.
Here simulations can help out \cite{gwynne-2016,gwynne-2017}.
Also, they are generally more time-efficient. Generating data such as trajectories is easy and often fully automated, while to this day, extracting trajectories from video footage remains a semi-manual process. Parameter variation is straight forward while it is costly and time-consuming in controlled experiments and impossible in field observations. Aside from practical and economic considerations, simulations can address situations that would be unethical in an experiment, \eg{} situations with dangerously high densities. 
Beside simulations, virtual and augmented reality environments offer possibilities to test such critical scenarios without endangering probands \cite{kinateder-2014b,feng-2018,lovreglio-2018}.

In addition, scientists can learn from studying the model, that is, the set of equations itself, provided the model matches reality sufficiently well. Whether a match is good or not depends on the particular research question.

\begin{figure}[h!]
	\centering
	\subfigure[Publications per year]{\includegraphics[height=0.4\linewidth]{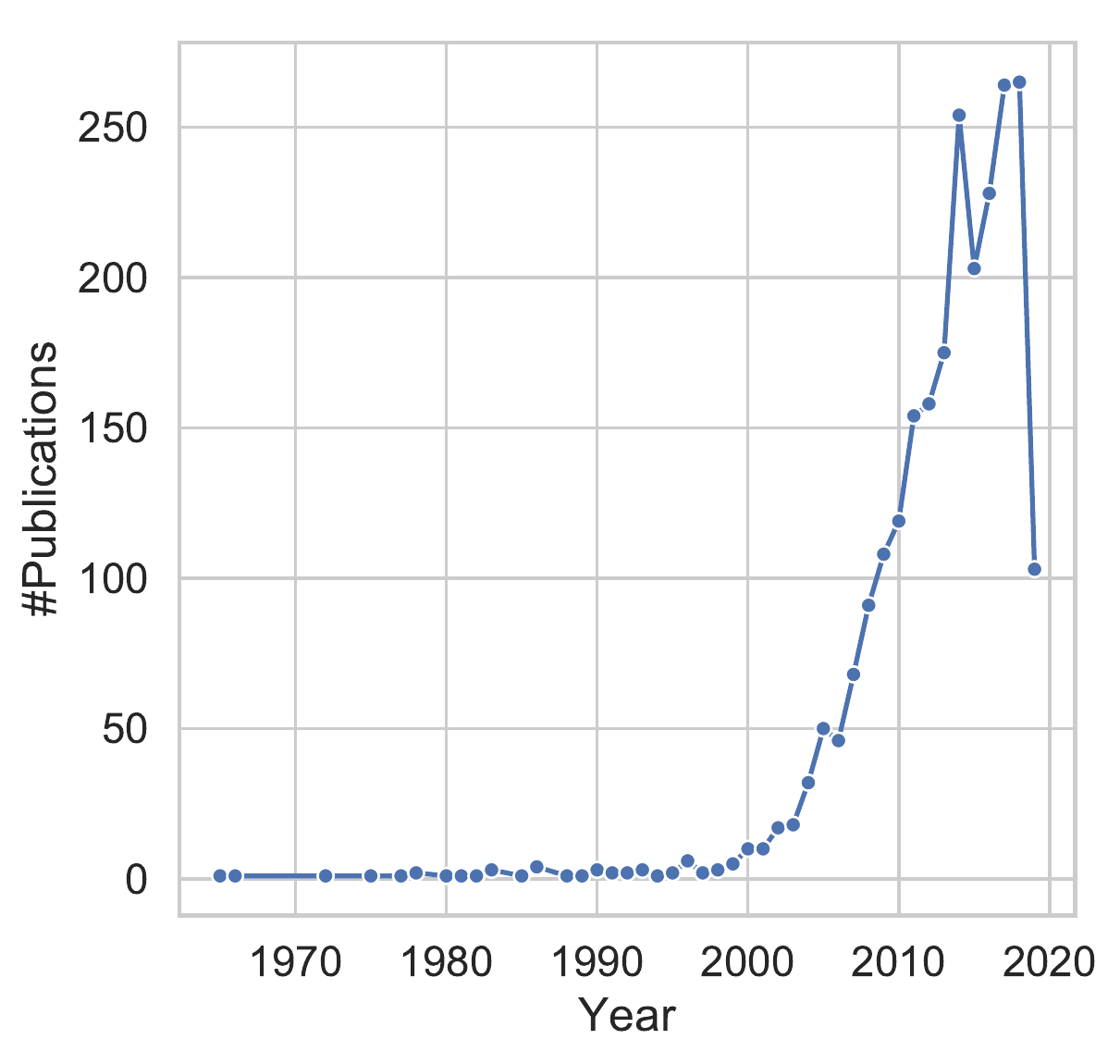}}
	\hfill
	\subfigure[Overall publications per subject]{\includegraphics[height=0.4\linewidth]{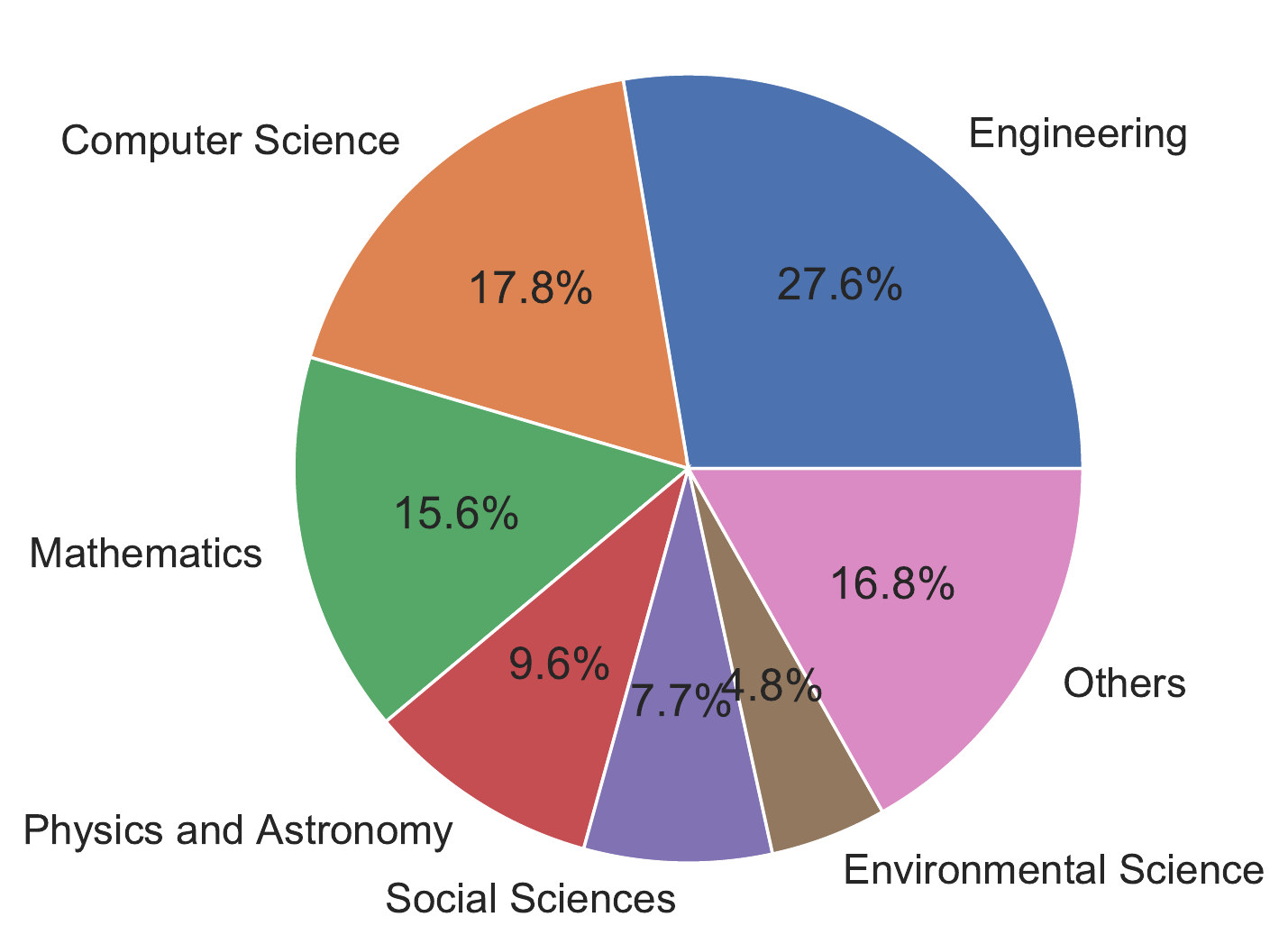}}
	\caption{Scopus search result for the term \enquote{pedestrian dynamics} on the 5th of June in 2019. Overall $2422$ document are listed.}
	\label{fig:PublicationsInPedestrianDynamics}
\end{figure}

At this point, researchers are faced with a challenge: the vast number of competing models from which they can choose \cite{wolfram-1984,helbing-1995,antonini-2005,seitz-2012}.
Pedestrian dynamics deal with the most sophisticated organism on earth, the human being. 
At the same time, the field is relatively new, with most of the progress and publications stemming from this millennium as illustrated by the search result shown in \cref{fig:PublicationsInPedestrianDynamics}. 
Therefore, it is not surprising that there is no universally accepted model of pedestrian dynamics. Indeed, the number of models and the level of detail they try to map is still growing.
 
One way to categorize pedestrian simulation models is by their scale: macroscopic and microscopic \cite{papadimitriou-2009}.
For both categories, the goal is to reproduce coarse-scaled quantities like densities (measured in persons per area) or flow (measured in persons per second).
While  macroscopic models, such as \cite{hughes-2001, treuille-2006}, see the crowd as a continuum, microscopic models see it as an aggregate of individuals.
Consequently, individuals are simulated based on their distinct attributes, such as their desired speeds.
In addition to coarse-scaled quantities, fine-scaled quantities like the velocity or step length of a pedestrian at a certain time can be measured.
Therefore, fine-scaled phenomena observed in the real world, such as the decrease in step length with increasing density, can be reproduced as well.

Microscopic models are often conceptually divided into three levels: (1) the strategic level: activity choice, (2) the tactical level: activity schedule, area and route choice to reach the area and (3) the operational level: walking behavior. A description can be found in \cite{hoogendoorn-2004}. The operational level, that is, the walking behavior, is embodied by a specific locomotion model.
%
In this contribution, we focus on the locomotion level. We argue that walking behavior must be modeled adequately before we can move on to modeling more complex situations, that is, before we can build a tactical and strategic level on top. Several approaches for modeling microscopic pedestrian dynamics have been developed in the last decades \cite{zheng-2009}. Among these are models based on cellular automata (CA) \cite{gipps-1985, burstedde-2001, kirchner-2003, was-2006, ezaki-2012, zhang-2012}, models based on ordinary differential equations (ODE) \cite{helbing-1995, chraibi-2010, chraibi-2011, dietrich-2014, tordeux-2014}, models based on cognitive heuristics \cite{seitz-2016c, xiao-2018} and models based on optimizing a utility, such as  the quasi-continuous optimal steps model \cite{seitz-2012, sivers-2014, sivers-2015, sivers-2016d}.

In short, even on the locomotion level alone, practitioners and researchers can pick and choose from a rich selection of solution proposals that may or may not be suitable for their research question. For example, in a large and open field scenario with millions of pedestrians a fast cellular automaton model may be a good, and perhaps the only choice, to gain first insights, while it will fail in situations with fine granularity. It is very difficult to make decisions on model descriptions alone. 

This is where we introduce \vadere{}. It is a free and open-source pedestrian simulation framework for microscopic pedestrian dynamics that was expressively conceived to facilitate comparison between locomotion models. Thus, it comes with pre-implemented versions of the most widely spread locomotion models. \vadere{} is released under the LGPL license.

Alternative implementations of the above-mentioned models are offered by a large number of commercial simulators, like PTV (social force model), LEGION (unknown model) and accu:rate (optimal steps model). However, to understand the dynamics of a simulation, not only the mathematical formulation of the model but also its implementation must be known and well understood. Indeed, the mathematical description will often differ from its implementation \cite{chraibi-2016}. 
For example, the mathematical model definition might be through an ordinary differential equation (ODE) as in force based models.
The algorithmic formulation, on the other hand, is through a numerical solver, such as Euler's method, which may -- or may not -- converge to the solution of the ODE. See \cite{koster-2013} for numerical issues.
Another example is the use of different pseudo random number generators with different starting seeds.
We assume a world of irrational numbers and true random numbers which do not exist in the coding world.
Those details are often not mentioned in the model description.
Thus, easy access to the code is a key requirement.

Several other open-source pedestrian dynamics simulators exist, for example FDS+Evac \cite{fdsevacontributors-2019}, JuPedSim \cite{chraibi-2016}, Menge \cite{curtis-2016}, MomenTUMv2 \cite{kielar-2016b} and SUMO \cite{sumocontributors-2015}. All frameworks share common conceptual characteristics but differ in their goals. 
\vadere{} is a lightweight simulator. 
The framework focuses on comparing existing and implementing new locomotion models and, at this point, solely covers the operational level, 
to keep the core simple and expandable. Developers can focus only on the model code, the rest is offered by the common code base. As a consequence, \vadere{} offers modern algorithms and data structures, and many more tools like a graphical user interface (GUI), Python scripts to compare simulation outputs and a command line interface to run multiple simulations in a systematic manner. All those features support and simplify the scientific work, especially the calibration and validation of locomotion models whose importance is emphasized in \cite{chraibi-2016}. 
In the spirit of interdisciplinary collaboration, the secondary focus of \vadere{} is its usability, which we continuously improve. The open-source approach allows other researchers to use and extend \vadere{} to their needs. 

In \cref{sec:MethodsAndMaterials}, we give an overview on existing locomotion models and pedestrian dynamics simulators that implement subsets of these models. In \cref{sec:ResultAndDiscussion}, we introduce the \vadere{} simulation framework. We show core features of \vadere{} which are relevant for users. Moreover, we reveal technical details which are essential when implementing a generic software framework that supports multiple locomotion models. In addition, we shed  light on how we test our simulation software while it is being improved and used continuously.

\section{Overview of locomotion models and simulation frameworks}
\label{sec:MethodsAndMaterials}

In this section, we cover two topics. Firstly, we introduce the three perhaps most widely used locomotion models which are all part of \vadere{}. Then, we give an overview on existing open-source crowd simulators that each support one or two of these locomotion models. From now on we refer to a simulated pedestrian as an agent while reserving the term \enquote{pedestrian} for the real world. 

\subsection{Locomotion models}
\label{sec:LocomotionModels}

In order to simulate realistic pedestrian behavior, adequate locomotion models are necessary. Researchers of various areas have been studied the way pedestrians move and have translated their findings into mathematical models. 

Historically, three development phases of locomotion models can be observed. In 1984, Stephen Wolfram introduced the idea of a cellular automaton to describe ``systems constructed from many identical components, each simple, but together capable of complex behavior" \cite{wolfram-1984}. 
Gipps and Marksj\"{o} picked up this idea to describe ``interactions between pedestrians" \cite{gipps-1985} through a cellular automaton. 

The second phase started with Helbing und Moln\'{a}r's publication of the social force model (SFM) in 1995 \cite{helbing-1995}. They renewed the idea of using  \enquote{forces} to describe pedestrian movements that was introduced by Hirai and Tarui in 1975 \cite{hirai-1975}. In the SFM, pedestrian motion is driven by a superposition of forces called \enquote{social} forces that, according to the authors, represent a ``measure for the internal motivation of the individuals to perform certain actions (movement)" \cite{helbing-1995}. The SFM was extended by several researchers such as \cite{chraibi-2010} to cope with some of its problems like the treatment of inertia. 

The current third phase started around 2000 when model development further branched out. ODE-based modeling turned from 
second order equations present in the SFM to first order equations, thus
eliminating several artifacts known in the SFM \cite{dietrich-2014b}. The results are so-called  velocity-based models \cite{dietrich-2014, tordeux-2014}. 
At the same time several more model types emerged: velocity-obstacle-based models \cite{fiorini-1998, berg-2011, curtis-2013} for collision avoidance inspired by robotics, rule-based models such as \cite{seitz-2016c, xiao-2018} and models which combine different approaches like the optimal steps model (OSM) \cite{seitz-2012}. Most models have been adjusted and extended to fit the findings from interdisciplinary experiments and studies. For example, the OSM was extended based on the (inter-) personal space theory \cite{sivers-2014, sivers-2015} and the social identity theory \cite{sivers-2016d}. 

Cognitive heuristics, social forces, Newtonian mechanics, velocity obstacles, social identities and personal space theory, computational geometry, queuing theory, cellular automata, and even fuzzy logic are ideas and techniques used in the development of modern microscopic pedestrian models. This shows that the exact nature of microscopic pedestrian behavior is open to debate. The scientific discourse will continue and the number of models will further increase. As a result, scientists need a software to compare them all.

In the following sections, we sum up the core ideas of three locomotion models which are widely used: cellular automata, forced based models and the optimal steps model.

\subsubsection{Common ground for all locomotion models}
\label{sec:CommonGroundForAllLocomotionModels}

\begin{figure}[t]
    \centering
    \includegraphics[width=\linewidth]{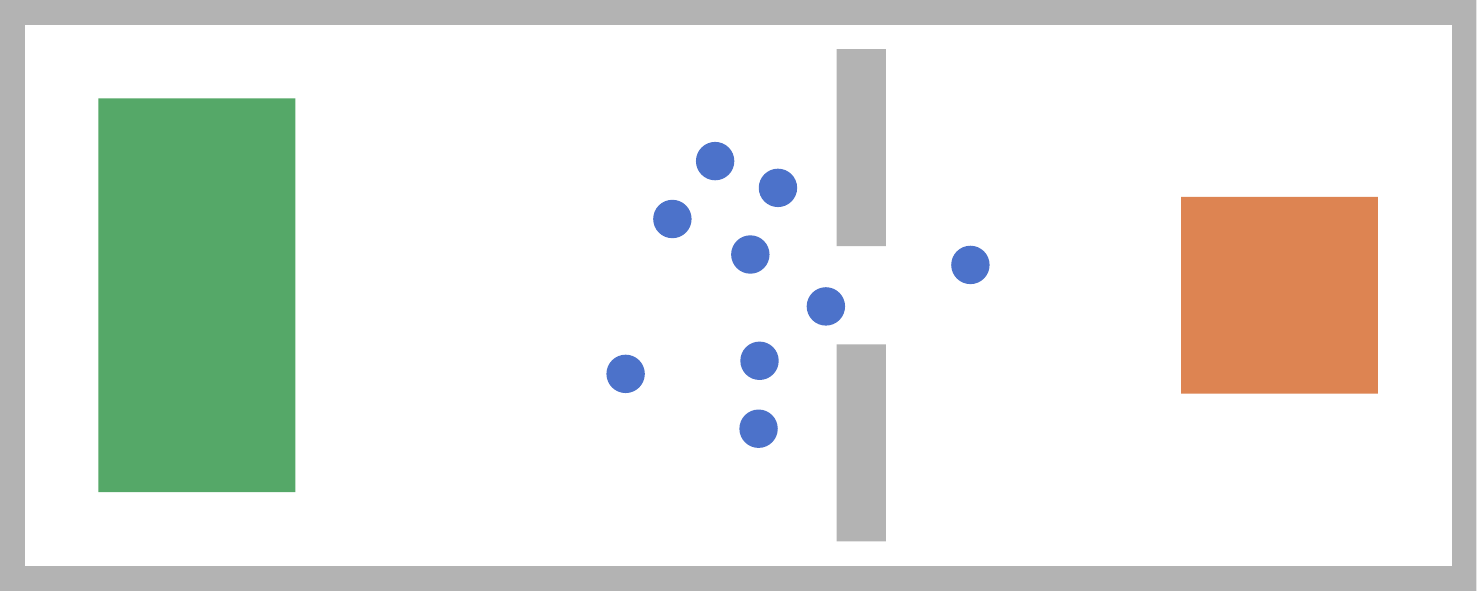}
    \caption{The four basic components of pedestrian crowd simulations: (1) agents (blue) who move from a (2) starting point (green) to a (3) destination (red) while avoiding (4) obstacles (gray).}
    \label{fig:SimulationsBasicComponents}
\end{figure}

\paragraph{Topography}
All locomotion approaches that we describe here use four basic modeling components: (1) agents --- simulated pedestrians --- who move from a (2) starting point (also called source or origin) to a (3) destination or target while avoiding (4) obstacles and other agents. These four basic components are illustrated in \cref{fig:SimulationsBasicComponents}. Together they form what we call topography.
\begin{figure}[h!]
	\centering
	\subfigure[Euclidean distance ignores obstacles.]{\includegraphics[width=0.40\linewidth]{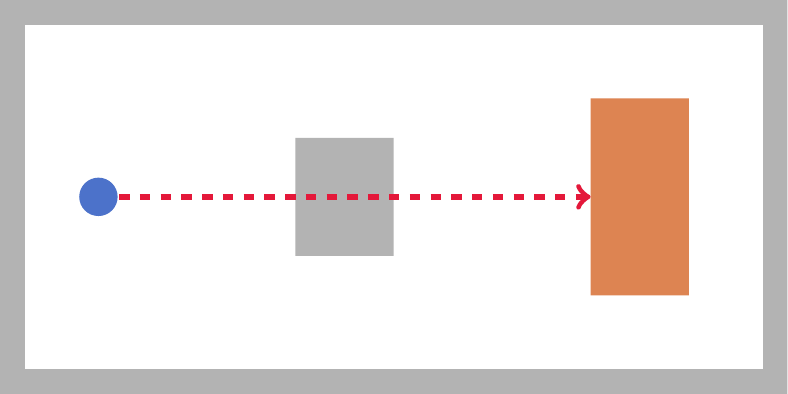}}
	\hfill
	\subfigure[Geodesic distance considers obstacles.]{\includegraphics[width=0.40\linewidth]{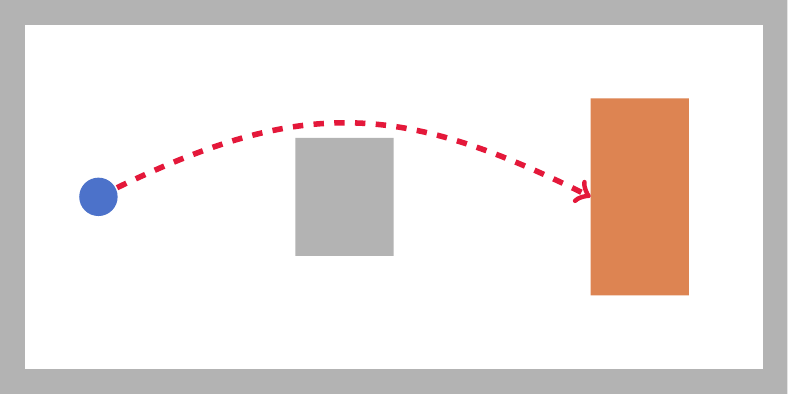}}
	\caption{Differences in pedestrian routing when using the Euclidean distance and the geodesic distance as a basis for calculating floor fields.}
	\label{fig:EuclideanVsGeodesicFloorField}
\end{figure}
\paragraph{Floor fields}
In addition, models use a floor field, also called scalar field \cite{seitz-2016b} or navigation field, which \enquote{guides} agents from an origin to a destination. These floor fields combine attraction and repulsion. While targets attract agents, obstacles and other agents repulse. 
In an alternative definition floor fields encode utility for
an agent modeled as ``homo economicus". Closeness to the destination increases utility, too close proximity to walls or other agents decreases utility. Floor fields can also be interpreted as a probability density function for the next move. The incorporation of the scalar fields depends on the model, but the general principle remains the same. 
%
The dominant part of most floor fields is the utility caused by proximity to the destination, or the attraction by the target. The simplest ansatz 
to encode proximity is to calculate the Euclidean distance from a destination to each point in the topography. The closer an agent is to the destination, the better its position. 
However, the Euclidean distance to the target fails whenever obstacles are located between an agent and its destination as illustrated in \cref{fig:EuclideanVsGeodesicFloorField}. Furthermore, it might fail even if the shortest path is free, for example, pedestrians do not necessarily follow the shortest path when passing doors \cite{lovreglio-2018}.
Therefore, in \vadere{}, we only use the geodesic distance to a target. The minimal geodesic path can be the shortest as well as the fastest path \cite{hartmann-2014b, koster-2014b}.  
\begin{figure}[h!]
	\centering
	\subfigure[$t = 0s$]{\includegraphics[width=0.47\linewidth]{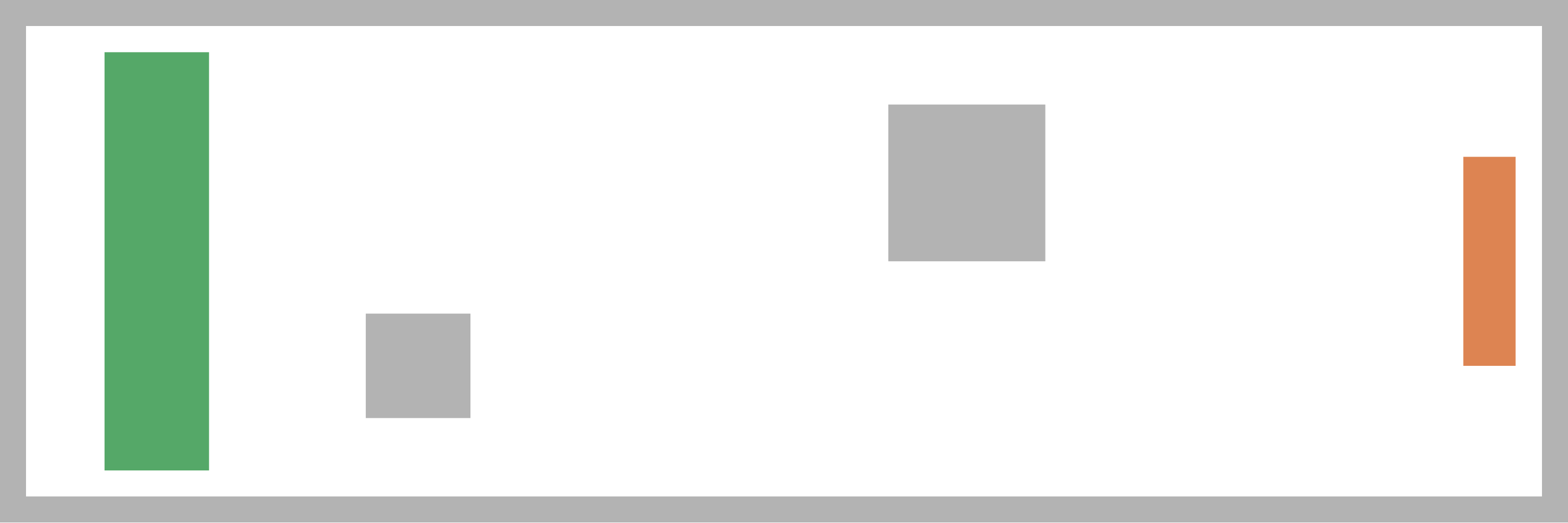}}
	\subfigure[Static floor field ($f = 1$).]{\includegraphics[width=0.47\linewidth]{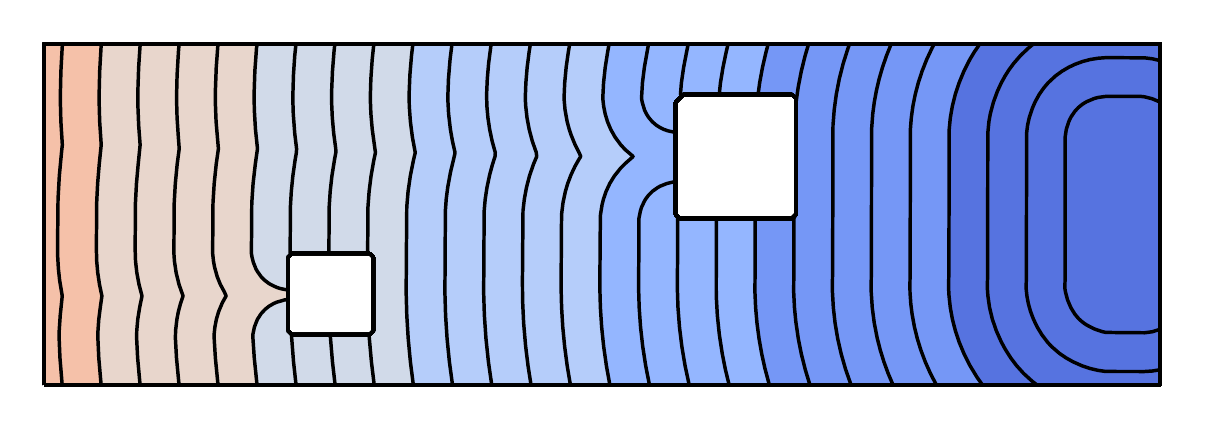}}
	\subfigure[$t = 6s$]{\includegraphics[width=0.47\linewidth]{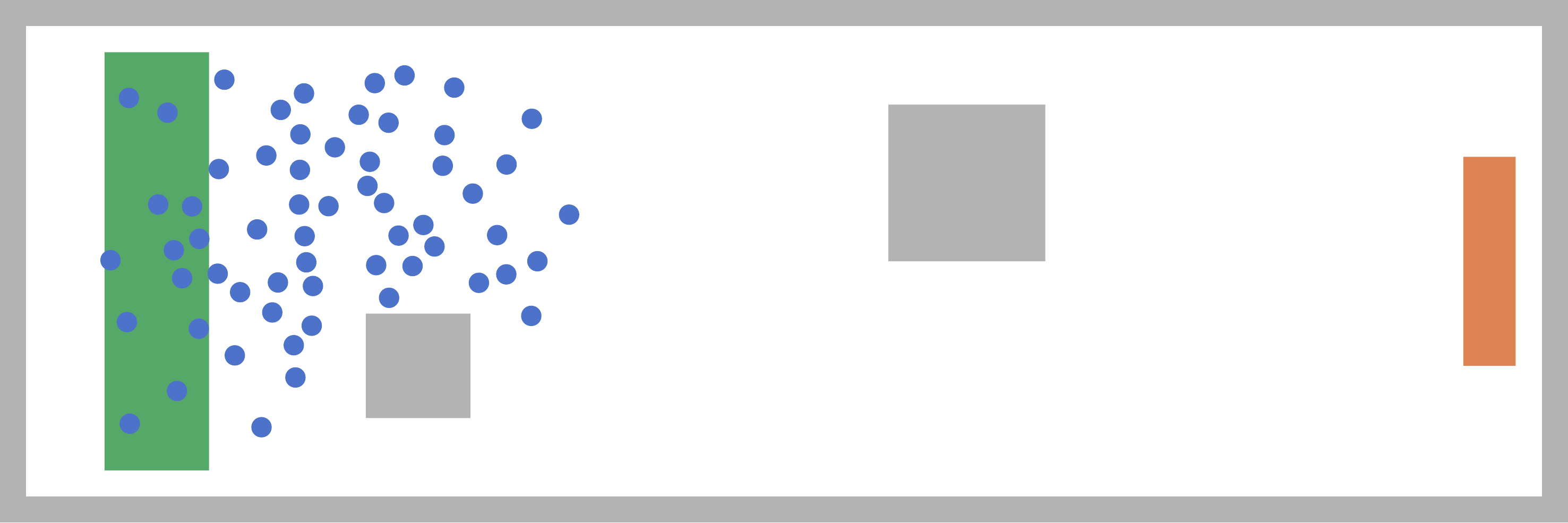}}
	\subfigure[Dynamic floor field, $t = 6s$]{\includegraphics[width=0.47\linewidth]{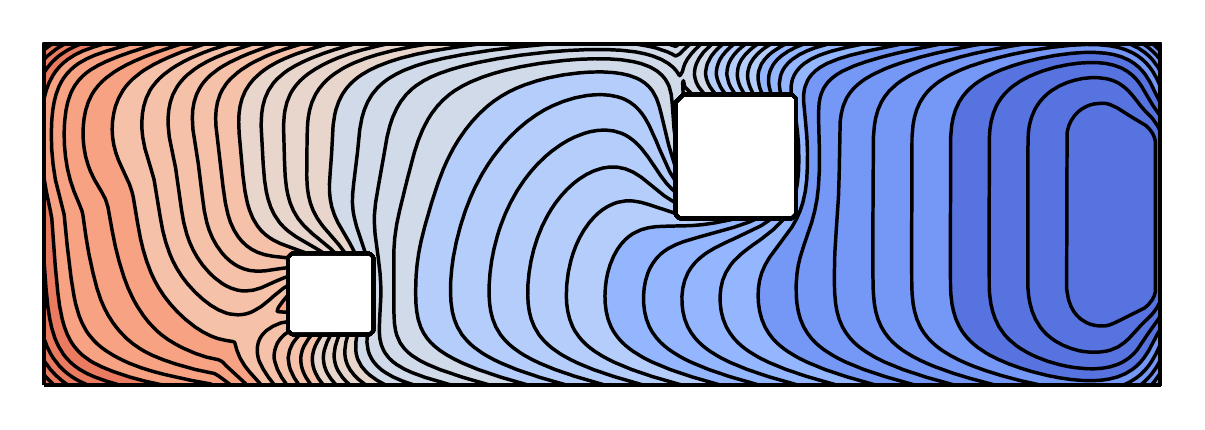}}
	\subfigure[$t = 32s$]{\includegraphics[width=0.47\linewidth]{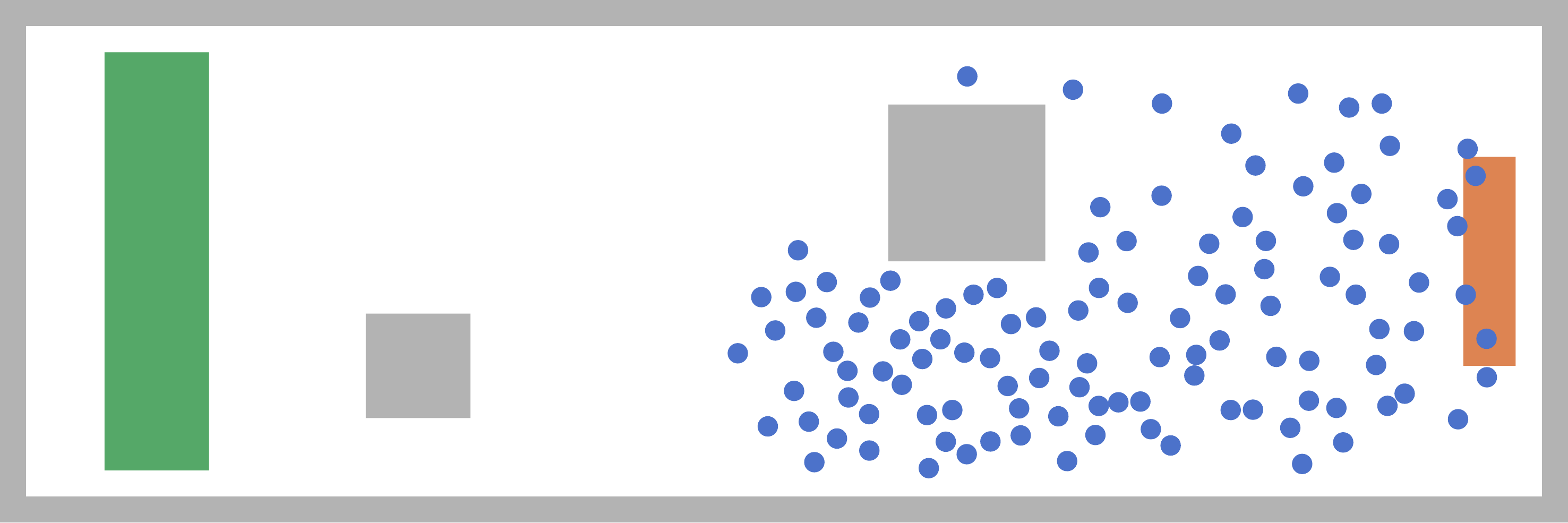}}
	\subfigure[Dynamic floor field, $t = 32s$]{\includegraphics[width=0.47\linewidth]{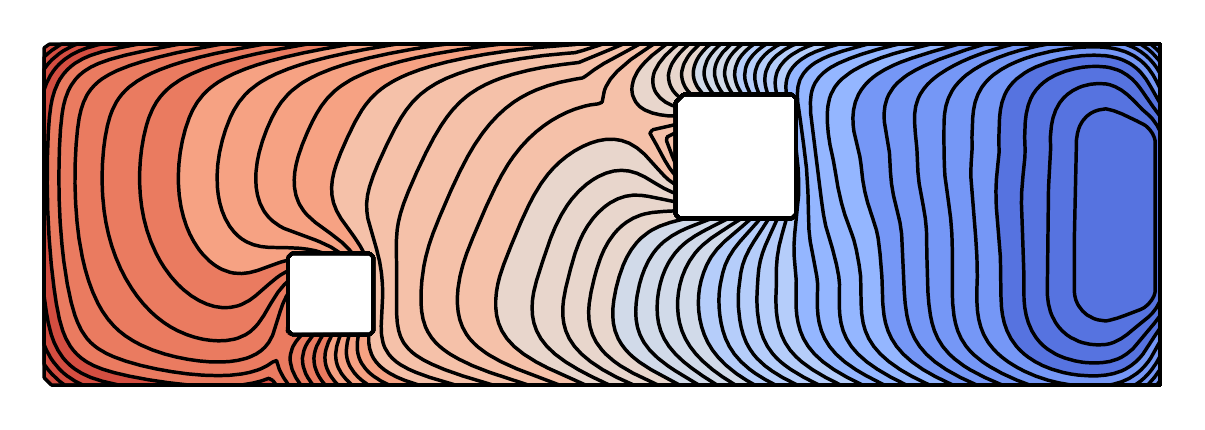}}
	\caption{Plot of the traveling time $u$. The wave propagates from the destination area (red) and flows around obstacles (gray). Dark blue colors represent small values \(u(x)\) while brighter blue and red colors represent higher values \(u(x)\). For a point $x$ inside an obstacle, the wave stops (speed $f(x) = 0$), \ie{} arrival time is infinite ($u(x) = \infty$). For a dynamic floor field $f$ depends on the current pedestrian and obstacle density. The situation is illustrated on the left and the plot on the right.}
	\label{fig:OSM-WaveFront}
\end{figure}
Solving the eikonal equation is a widely used approach to obtain the geodesic distance between a destination and each position in the topography. The eikonal equation is a non-linear partial differential equation:
\begin{equation}
\begin{split}
\vert \vert \nabla u(x) \vert \vert f(x) &= 1\quad\mbox{for}\ x \in \Omega \subset \mathbb{R}^2\\
u(x) &= 0\quad\mbox{if}\ x \in \Gamma.
\label{eq:EikonalEquation}
\end{split}
\end{equation}
The solution $u$ of the eikonal equation is the travel time of a wave front that starts at the target region $\Gamma \subset \Omega$ and propagates across the topography area $\Omega$ with speed $f \geq 0$. For all positions inside obstacles, $f$ is zero, that is, the wave stops. Compare to \cref{fig:OSM-WaveFront}. If $f = 1$ everywhere else, $u(x)$ is the shortest distance to the target area $\Gamma$ starting at position $x$ while skirting obstacles. To achieve more realistic medium scale avoidance behavior, one can choose a speed function $f$ for the wave front that depends on the pedestrian density. 
By lowering $f$ at crowded locations, that is slowing the wave down, medium-scale avoidance of other pedestrians can be modeled \cite{hartmann-2014b, koster-2014b}. In this case, the floor field becomes dynamic and must be recomputed regularly.
\subsubsection{Cellular automata (CA)}
\label{sec:CellularAutomata}
\begin{figure}[!h]
    \centering
    \includegraphics[width=\linewidth]{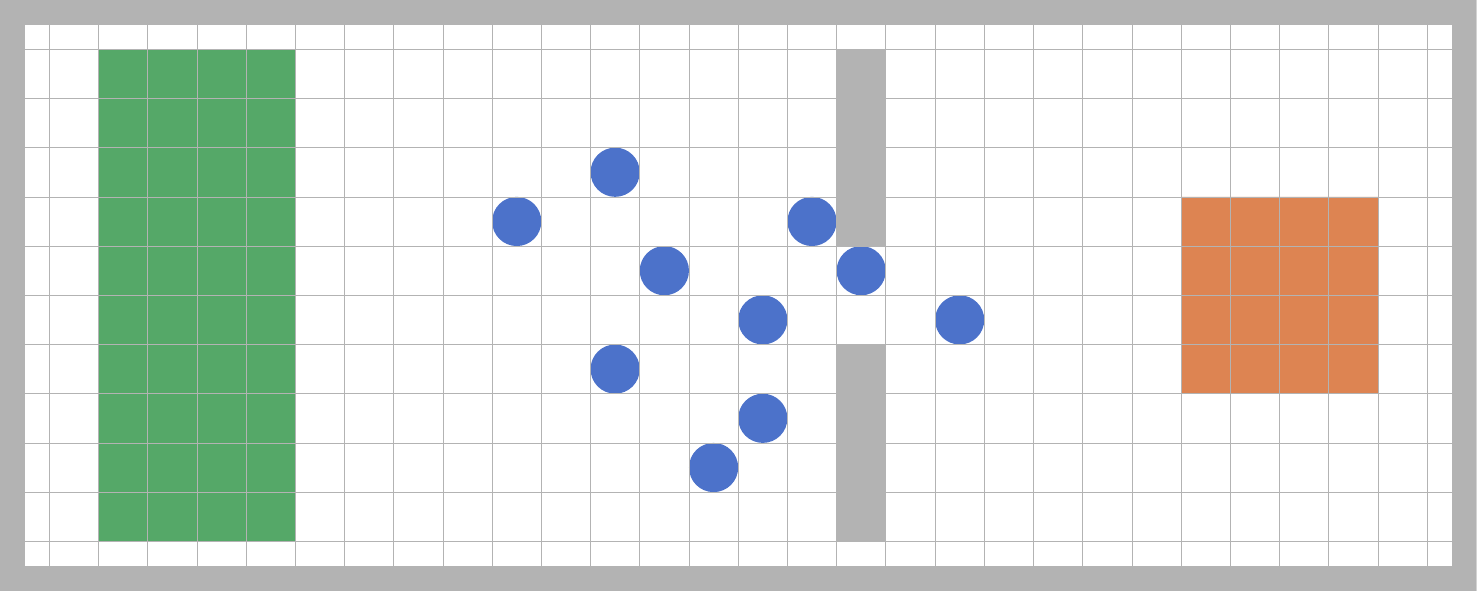}
    \caption{Cellular automaton based on an evenly-spaced grid over the topography content. Agents move from a source (green) green area to a target (red).}
    \label{fig:CellularAutomata-Grid}
\end{figure}

Stephen Wolfram generalized the idea of a cellular automaton:

\begin{quotation}
	Cellular automata are mathematical idealizations of
	physical systems in which space and time are discrete,
	and physical quantities take on a finite set of discrete
	values. A cellular automaton consists of a regular uniform lattice (or \enquote{array}), usually infinite in extent, with a
	discrete variable at each site (\enquote{cell}). [...] A cellular automaton evolves in discrete time steps, with the value of the variable at one site being affected by the values of variables at sites in its \enquote{neighborhood} on the previous time step.
	\cite{wolfram-1983}
\end{quotation}
In the context of pedestrian dynamics, the concept of CA is implemented as follows. The topography area is divided into cells of equal shape that are either empty or occupied by a pedestrian, a target or an obstacle. The state of a cell is updated at each time step, that is, agents move from cell to cell according to certain rules.  
In most CA-models for pedestrian dynamics the rules depend on a floor field as described above and positions are updated instantaneously. 
There are several ways to define the update order of pedestrians. 
The order in which pedestrians are updated has an impact on the simulation outcome \cite{seitz-2014b}. It is rarely parallel, to avoid collisions, and thus differs from Wolfram's original definition of cellular automata. Furthermore, the state of a cell is not purely defined by the state of neighboring cells but also by some global information. \cref{fig:CellularAutomata-Grid} visualizes the concept of a cellular automaton at one specific time step.

\begin{figure}[!h]
    \centering
    \subfigure[Movement artifacts: Agents walk in a zig-zag manner. ]{\includegraphics[width=0.45\linewidth]{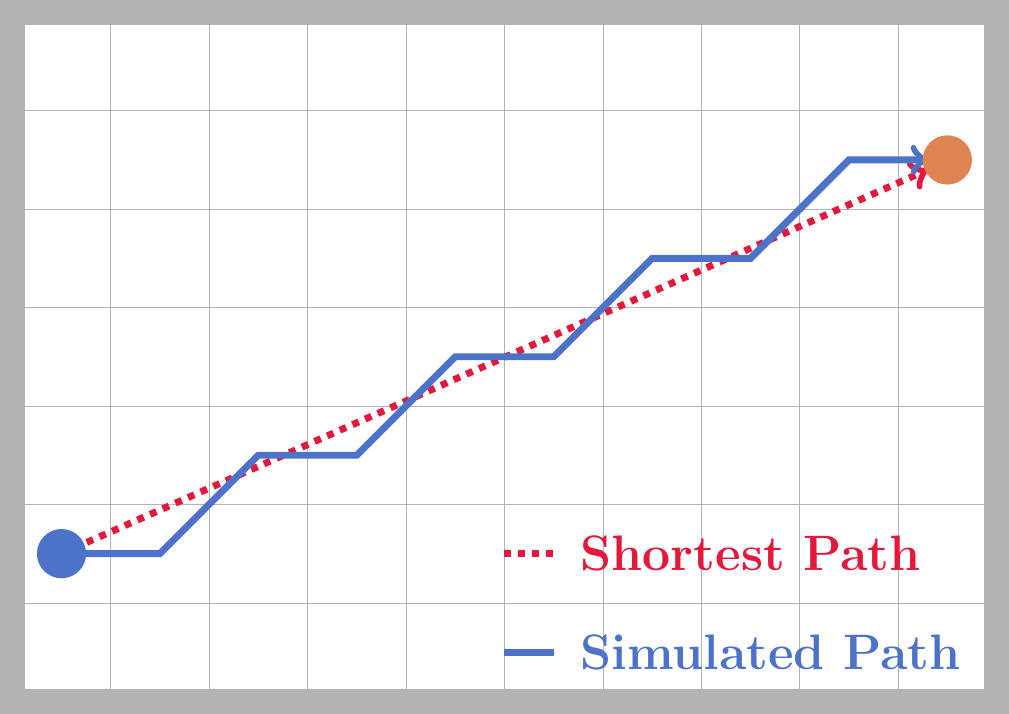}}\qquad
    \subfigure[Impact of the grid resolution: A too coarse grid hinders agents to pass through a narrow passage.
    ]{\includegraphics[width=0.45\linewidth]{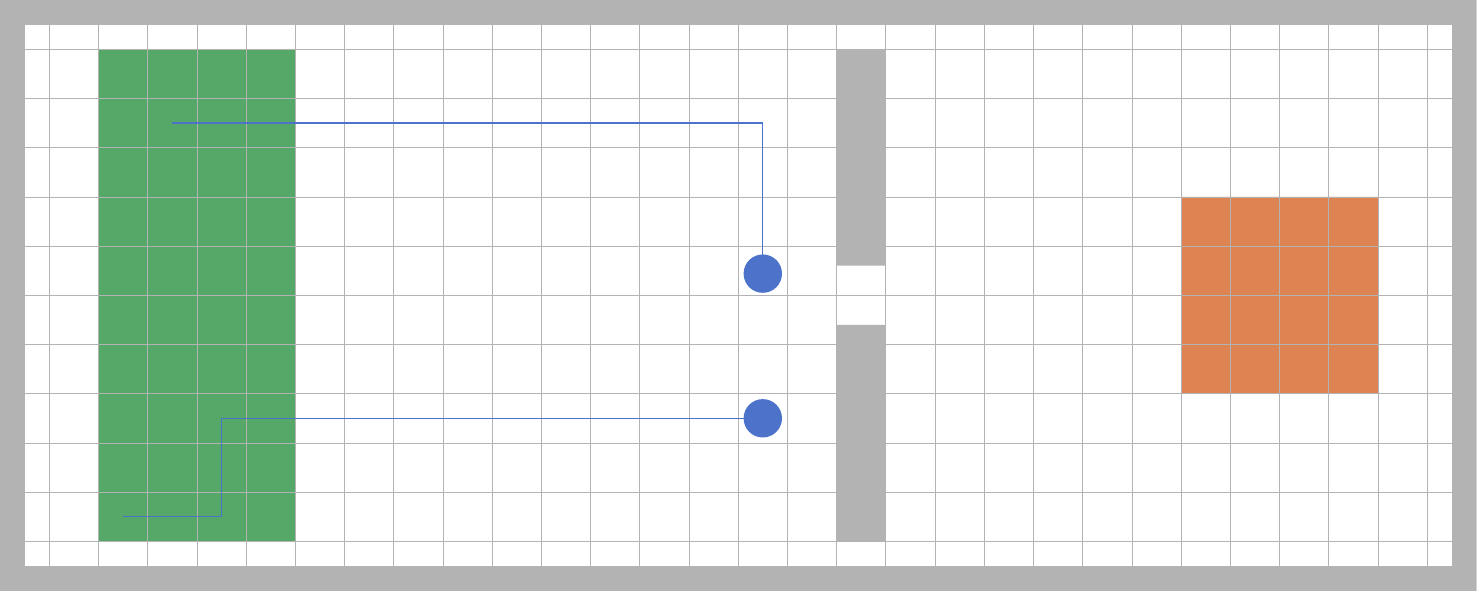}}
    \caption{Two drawbacks when using cellular automata: movement artifacts and impact of the grid resolution on the motion.}
    \label{fig:CAsDrawbacks}
\end{figure}

The major advantage of cellular automata is their simplicity with respect to understanding, implementation and computation. Compared to other locomotion models like ODE-based models, which require integration, cellular automata are significantly less computationally expensive. 
Thus, CA-models were more popular in the past when computational power was a more limited resource. However, the coarse discretization of cellular automata leads to artifacts. More complex cellular automata 
where each agent is contained in several small cells
attain a small variation of step lengths and directions \cite{was-2014} but not the full range. 
Additionally, agents tend to move along zig-zag trajectories because of the grid structure and may get stuck in narrow passages that are not resolved by the grid.
\cref{fig:CAsDrawbacks} visualizes these artifacts.  

\vadere{} implements cellular automata as a special case of the optimal step model  where certain constraints hold. See \cref{sec:OptimalStepsModel}. Thus, the \vadere{} implementation does not do justice to the computational efficiency a cellular automaton can achieve. However, it allows scientists to judge for themselves
whether or not the grid artifacts would compromise the validity of their simulation study.
\subsubsection{Force-based models}
\label{sec:SocialForceModel}

In 1975 a first force-based model, inspired by the motion of a shoal, was introduced in Japan \cite{hirai-1975}. Four years later, \cite{okazaki-1979} proposed a model inspired by forces between magnets. 
The best known and analyzed force-based model is Helbing's and Moln\'{a}r's
social force model (SFM) from 1995 \cite{helbing-1995}. 
Extensions were proposed by Helbing and his group \cite{johansson-2007, moussaid-2010} as well as  others \cite{yu-2005, lakoba-2005, pelechano-2007, parisi-2009, chraibi-2010, koster-2013}.
Agents are treated like particles that are accelerated or decelerated by  \enquote{social} forces taking Newton's second law of dynamics as a guiding principle. The primary force $F_l^t$ of all these models is the force that drives an agent $l$ towards its destination
\begin{equation} \label{eq:targetForce}
	F_l^t = \frac{1}{\tau_l} \left(v_l^0 e_l - \dot{x}_l \right).
\end{equation}
Here $v_l^0$ is the agent's free-flow velocity, $x_l$ its position, $e_l$ the direction pointing towards the target region and $\tau_l$ the agent's reaction time. If the agent can move unhindered, that is, if there are no obstacles or other agents, this is the only force affecting the agent.
In the following, we describe the SFM in its original form without its extensions. 

While obstacles always evoke a repulsive force pushing agents away, forces defined between two agents can either attract, \eg{} to model groups, or repulse. Acceleration is described through 
\begin{equation}
	\dot{w_l} = F_l + fluctuation,
\end{equation}
where $w_l$ is the unrestricted velocity induced by the sum of all forces $F_l$ acting on agent $l$ \cite{helbing-1995}. The $fluctuation$ term was added to take random variations into account. The actual velocity $\dot{x}_l$ of agent $l$ is limited to some maximum speed $v^{max}_l$:
\begin{equation}
	\dot{x}_l = v_l(w_l) = 	\begin{cases}
	w_l & \text{if } \vert\vert w_l \vert\vert \leq v_l^{max} \\
	w_l  \frac{v_l^{max}}{\vert\vert w_l \vert\vert} & \text{otherwise}.
	\end{cases}
\end{equation}
In its original version, the direction towards the target $e_l$ is the normed vector pointing to the closest point of the target region. This entails numerical problems when an agent \enquote{steps} on the target \cite{koster-2013}. Instead the normed gradient of $u(x_l)$ at the agent's position $x_l$, defined by \cref{eq:EikonalEquation}, that is
\begin{equation}
e_l = \frac{-\nabla u(x_l)}{\vert\vert\nabla u(x_l)\vert\vert},
\end{equation}
gives a more sophisticated definition of the target direction \cite{kretz-2011} which is used in \vadere{}.

The social force model has been reported to reproduce collective phenomena like lane formation, the ``faster-is-slower" effect, oscillations at bottlenecks and clogging at exit doors \cite{chraibi-2010}. Johansson et al.\,provide an overview of extensions of the social force model, such as the inclusion of physical forces, a constrained interaction range and limiting the number of other pedestrians considered \cite{johansson-2014}.
The social force model is very popular, but it has also been criticized for
artifacts that stem from its proximity to Newtonian mechanics, namely 
oscillating trajectories and overlapping agents \cite{chraibi-2014, dietrich-2014b}, and for numerical issues \cite{koster-2013}. Another problem of all force-based models arises with the multiple roles of the relaxation time $\tau_l$ in \cref{eq:targetForce}. It affects how precisely agents follow their preferred path, and at the same time, how they avoid collisions \cite{johansson-2014}. These two behaviors may not be correlated.
The authors believe that it is up to the scientists to decide which artifacts can be tolerated in view of their specific research questions. \vadere{} facilitates this decision: scientists can visually compare simulation outcomes for test scenarios of their choosing.

\subsubsection{Optimal steps model (OSM)}
\label{sec:OptimalStepsModel}
The optimal steps model (OSM) defines motion by a series of discrete footsteps. Like cellular automata-based but unlike ODE-based models, each footstep is modeled explicitly as a discrete time event. It starts and ends at a specific time but is performed instantaneously. The model was introduced in \cite{seitz-2012} and enhanced in \cite{sivers-2014, sivers-2015, sivers-2016d}. In this contribution we refer to the latest version of the OSM described in \cite{sivers-2016d}. As many other models the OSM seeks a balance of goals: reach a target while avoiding obstacles and other agents.   

 \begin{figure}[b]
	\centering
	\subfigure[Slowing down due to smaller footsteps.]{\includegraphics[width=0.49\linewidth]{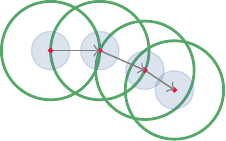} \label{fig:OSMslowingDown}} 
	\hfill
	\subfigure[A sample of possible next positions (red).]{\hspace{0.10\linewidth}\includegraphics[width=0.25\linewidth]{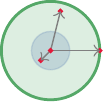}\hspace{0.10\linewidth}\label{fig:OSMNextStep}}
	\caption{Illustration of footsteps of an agent in the OSM. The circles (green) indicate the actual step radius and the area (blue) represents the agent torso. Since each step takes the same amount of time, smaller steps (left) lead to a lower walking velocity. All positions inside the circles (green) are possible next positions.}
	\label{fig:OSMStepping}
\end{figure}

During the simulation, each agent $l$ performs a series of steps as depicted in \cref{fig:OSMslowingDown}. The next foot position is found by
optimizing utility within a circle around the agent. The utility is encoded in a floor field $P_l$ as described in \cref{sec:CommonGroundForAllLocomotionModels}. 
$P_l$ can also be interpreted as a potential field.
The radius $s^0_l$ of the step circle is correlated with the agent's free-flow velocity $v^0_l$. More precisely
 \begin{equation}
 s^0_l = \beta_0 + \beta_1 \cdot v^0_l + \epsilon, \quad \beta_0, \beta_1 > 0, \ \epsilon \sim \mathcal{N}(0,\sigma^{2}), \label{eq:SpeedStrideLength}
 \end{equation}
 where the error term $\epsilon$ is assumed to be normally distributed \cite{seitz-2012}. In crowded situations the agents make shorter step, shorter than $s^0_l$, while the duration between steps, $s^0_l / v^0_l$, remains unchanged. Thus deceleration in a crowd is an emergent behavior of the OSM. 
\vadere{}'s OSM uses an event driven update scheme \cite{conway-1959} and each step of an agent is an event.
Stepping events of multiple agents are ordered in an event queue and executed accordingly \cite{seitz-2014b}. 

The potential field $P_l$ of an agent $l$ is given by a sum of sub-utilities or sub-potentials: $P_t$ contributes attraction to the target or utility from proximity to the target and is given by the solution of the eikonal \cref{eq:EikonalEquation} for all agents that share a target. It is defined globally, that is, on the whole simulation area.
$P_{p,l,i}$ causes local repulsion (or dips in utility) by other agents, and  $P_{o,j}$ local repulsion from obstacles:
\begin{align} \label{eq:OSMAggregated}
	P_l(x) &= P_{t,l}(x) + P_{p,l}(x) + P_{o}(x) \\
	P_{p,l}(x) &= \sum\limits_{i=1, i \neq l}^n P_{p,l,i}(x) \\
	P_o(x) &= \max\limits_{j \in \{1, \ldots, m\}} P_{o,j}(x),
\end{align}
with $n$ agents and $m$ obstacles in the topography (compare \cite{seitz-2012} and \cite[p. 63]{seitz-2016}).

The agent potential $P_{p,l,i}$ translates the theory of (inter-) personal space \cite{hall-1990} into a mathematical formula. 
$P_{p,l,i}$ is the sum of three functions: $p^{\text{per}}_{l,i}$, $p^{\text{int}}_{l,i}$ and $p^{\text{tor}}_{l,i}$ which correspond to the personal space, the intimate space and the torsos of agents $l$ and $i$. 
\begin{equation}
	P_{p,l,i}(x) = 	\begin{cases}
	P_{p,l,i}^{\text{tor}}(x)
	 & \text{if } d_i(x) < r_i + r_l  \\
	P_{p,l,i}^{\text{int}}(x) & \text{if } r_i + r_l \leq d_i(x) < \delta^{\text{int}}_l \\
	P_{p,l,i}^{\text{per}}(x) & \text{if } \delta^{\text{int}}_l \leq d_i(x) < \delta^{\text{per}}_l \\
	0 & \text{else}.
	\end{cases}
\end{equation}
with
\begin{align}
	P_{p,l,i}^{\text{tor}}(x) &= p^{\text{per}}_{l,i}(x) + p^{\text{int}}_{l,i}(x) + p^{\text{tor}}_{l,i}(x) \\
	P_{p,l,i}^{\text{int}}(x) &= p^{\text{per}}_{l,i}(x) + p^{\text{int}}_{l,i}(x) \\
	P_{p,l,i}^{\text{per}}(x) &= p^{\text{per}}_{l,i}(x)
\end{align}
Here $y_i$ is the position, $r_i$ the radius of the torso of agent $i$ and $\delta^{\text{per}}_l$, $\delta^{\text{int}}_l$ are the radii of the personal and intimate space of agent $l$. $d_i(x)$ gives the Euclidean distance between $y_i$ and $x$, that is,
\begin{equation}
d_i(x) = \vert\vert y_i - x\vert\vert.
\end{equation}
In the current implementation of the OSM the personal spaces and radii of all agents are equal. For more details we refer to \cite{sivers-2015, sivers-2016b}.

Let $x_{k}$ be the position of an agent $l$, then its next position $x_{k+1}$ minimizes $P_l$ (or maximizes $-P_l$) with respect to all reachable positions, that is,
\begin{equation}
	x_{k+1} = \argmin\limits_{y \in \disc(x_k)} P_l(y)
\end{equation}
with 
\begin{equation}
	\disc(x_k) = \{ y \in \mathbb{R}^2 : \Vert y - x_k \Vert  \leq s^0_l \}.
\end{equation}
Different floor field definitions  can be used to reproduce  competitive jostling or more cooperative queuing at bottlenecks \cite{zoennchen-2013}
and several other behaviors 
\cite{sivers-2015, seitz-2015, koster-2015b, koster-2014b}. The optimization is robust and with carefully designed potentials, the OSM is free  of overlaps and oscillations.

\begin{figure}[h]
	\centering
	\includegraphics[width=0.35\linewidth]{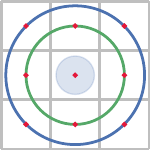}
	\caption{Illustration of the OSM mimicking a cellular automaton. 
		If the agents can only reach the points (red) on the circle (green), we obtain a cellular automaton considering the von Neumann neighborhood. Including also the points on the larger circle (blue) expands the neighborhood to the Moore neighborhood.}
	\label{fig:OSMToCA}
\end{figure}

Using the OSM leads to solving many optimization problems. 
For each step of each agent a non-trivial optimization problem has to be solved. 
This optimization is computational expensive and requires most of the overall computation time. 
Introducing more complex potential functions complicates the evaluation of $P_l$ which contributes directly to the computation time of the optimization. 
This is aggravated by fact that a strict event-driven update hinders parallelization of the computation. 
Therefore, simulating thousands of agents in real-time using the OSM with Vadere is not yet possible. 
Performance improvement is one of the topics we are currently working on.
\paragraph{Mimicking cellular automata with the optimal steps model}
The OSM can mimic motion in a cellular automaton by restricting the optimization to equidistant points on a circle \cite{seitz-2012}. See \cref{fig:OSMToCA}.
\subsubsection{Other locomotion models}
\label{sec:OtherLocomotionModels}
\begin{figure}[!h]
    \centering
    \includegraphics[width=\linewidth]{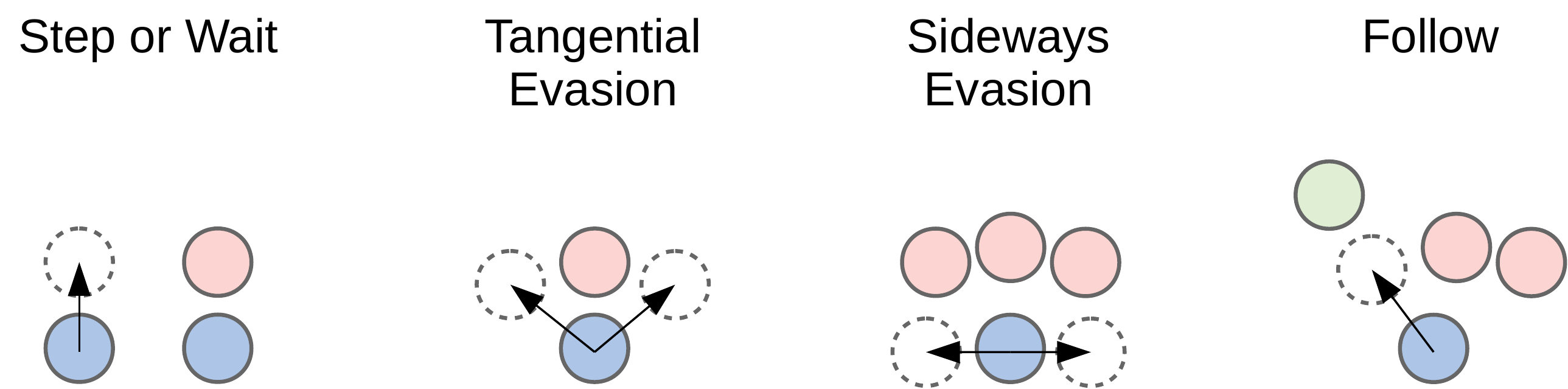}
    \caption{Illustration of behaviors with the four heuristics defined by the behavioral heuristics model: step or wait, tangential evasion, sideways evasion and follow (compare \cite{seitz-2016c}).}
    \label{fig:BHM-Seitz2016}
\end{figure}

We present a short description of other models that are implemented in \vadere{}. For  details we refer to the respective publications.
\paragraph{Gradient navigation model (GNM)}
The gradient navigation model is similar to ve\-lo\-city-based models that avoid the acceleration equation of the harmonic oscillator  characteristic for force-based models.
The GNM uses a superposition of gradients of distance functions to directly change the direction of the velocity vector \cite{dietrich-2014}. Speed is a scalar that is adapted according to the difference between the desired speed and the actual speed while considering that reaction is delayed.
Therefore, the relaxation time $\tau$ is disentangled from path finding.
\paragraph{Behavioral heuristics model (BHM)}
The behavioral heuristics model (BHM) follows a completely different approach \cite{seitz-2016c}: simple heuristics rooted in cognitive psychology determine each agent's next step. Each agent uses one of four heuristics that correspond to four levels of cognitive capacity. 
The simplest is the step or wait heuristic: The agent checks whether the next full step in the desired direction leads to a collision. 
If so, it waits and if not, it steps forward. The tangential evasion heuristic allows the agent to 
evade another agent in its path tangentially, if the evasion step does not cause a collision. This requires two additional collision tests. Two more levels of cognitive effort are modeled. See \cref{fig:BHM-Seitz2016}.   
%
\paragraph{Steering behaviors model}
Reynolds' steering behavior model offers simulation as an alternative to scripting the path of each individual to reproduce the aggregated motion of a flock of birds, a herd of land animals, or a shoal of fish \cite{reynolds-1987}.
The model defines a set of steering behaviors which can be combined to establish more complex goals such as walking through a narrow corridor.
Steering behaviors are motivated by the visual beauty observed in nature. 
Reynolds' goal was to reproduce realistic looking motion pattern for animation and games rather than studying pedestrian behavior.
Due to his target group the computational complexity of the model is low.
Furthermore, the ability to combine different steering behaviors and the code accessibility furthers the software development process.
New steering behaviors can be introduced without inventing a whole new model and an open-source implementation of Reynolds' work is available on his website.
Despite the different target audience, Reynolds' model is used in pedestrian dynamics to compare it against other models. 
In \cite{wolinski-2014}, for example, the authors compare the social force model, a velocity-obstacle-based model presented in \cite{berg-2011} and Reynolds' steering behavior model.

\subsection{Existing simulation frameworks}
\label{sec:ExistingSimulators}
Several companies like, PTV, LEGION and accu:rate, offer licenses for commercial crowd simulators based on microscopic models. However, their implementations are proprietary. The code cannot be analyzed to understand it in detail and to ensure that computer experiments can be reproduced exactly. 
This hinders comparison between models and thus the knowledge transfer between researchers that we strive for in this contribution. 
Therefore, we limit this discussion to open-source frameworks.
The overview in \cref{tbl:ExistingOpenSourceSimulators} lists frameworks that have been documented through publications or tutorials and have undergone recent development activities. 
The table shows the initial release date, the programming language the simulator is based on, the number of files and the lines of code \footnote{Lines of code excluding test code, blank lines and comments counted with \enquote{cloc} software tool. See \cref{app:ObtainLOCsForDifferentSimulators} for more details about how lines of code were obtained.}.
For a more extensive, but never exhaustive overview, we refer to \cite[p. 17--22]{sivers-2016b}.
%
%

Among the six simulators in \cref{tbl:ExistingOpenSourceSimulators}, SUMO (simulation of urban mobility) plays a special role. 
SUMO focuses on complete intermodal traffic systems including road vehicles, public transport and pedestrians, while FDS+Evac, JuPedSim, Menge and MomenTUMv2 concentrate on the actual pedestrian dynamics.
\begin{table}[h!]
    \centering
    {\scriptsize
    \begin{tabular}{lccrr}
        \textbf{Simulator Name} & \textbf{Released} & \textbf{Language} & \textbf{Files} & \textbf{LoC}\\
        \toprule
        FDS+Evac \cite{korhonen-2007} & 2007 & Fortran & 715 & 249,702 \\
        JuPedSim \cite{chraibi-2016} & 2014 & C++ & 774 & 173,973 \\
        Menge \cite{curtis-2016} &  2014 & C++ & 697 & 67,476 \\
        MomenTUMv2 \cite{kielar-2016} & 2016 & Java & 814 & 56,569 \\
        SUMO \cite{krajzewicz-2012} & 2001 & C++ & 1,618 & 253,472 \\
        \vadere{} & 2010 & Java & 977 & 73,145 \\
        \bottomrule
    \end{tabular}
    \caption{Open-source simulation frameworks for pedestrian dynamics (in alphabetical order).}
    \label{tbl:ExistingOpenSourceSimulators}
}
\end{table}
\paragraph{FDS+Evac}
\label{sec:FDSEvac}
The Fire Dynamics Simulator (FDS) has been developed by the National Institute of Standards and Technology (NIST) since 2000 \cite{mcgrattan-2019}. FDS started as a pure large-eddy simulator for slow flows which focuses on smoke and heat transport from fires. In the following years, the VTT Technical Research Centre of Finland joined this development and integrated the evacuation module FDS+Evac into FDS in 2007. FDS+Evac focuses on simulating human egress.
The simulation framework consists of four components: \textbf{(1)} The simulation core which is called FDS. \textbf{(2)} The graphical user interface Smokeview (SMV) that is used to display the output of FDS. \textbf{(3)} The FDS+Evac submodule for FDS to integrate agent-based simulations of humans and \textbf{(4)} additional third-party tools for visualization, pre- and post-processing. FDS+Evac uses the social force model \cite{helbing-1995} to move agents in a 2-dimensional plane and offers grouping behavior and different exit selection strategies for agents \cite{korhonen-2007}.
\paragraph{JuPedSim}
\label{sec:JuPedSim}
JuPedSim's development is mainly driven by the Forschungszentrum J\"{u}lich.
JuPedSim is a framework for the simulation of pedestrian dynamics at a microscopic level that focuses on evacuation scenarios. 
JuPedSim consists of four modules: \textbf{(1)} JPScore simulates the movement of agents. JPScore provides three models on the tactical layer: a shortest path strategy, a quickest path strategy and a cognitive map to explore the environment, \eg{} to discover doors. On the operational layer, JPScore provides three continuous models based on ordinary differentials: 
the force-based generalized centrifugal force model \cite{chraibi-2010}, the collision-free velocity model \cite{tordeux-2015b} and the wall-avoidance model \cite{graf-2015}. \textbf{(2)} JPSreport includes tools for density, velocity and flow measurements to analyze agent trajectories. \textbf{(3)} JPSvis visualizes simulation results through 2D or 3D animations. JPSvis can be directly connected to JPScore to get an online visualization of a simulation run. \textbf{(4)} JPSeditor is a tool for editing model parameters and the topography.

\paragraph{Menge}
\label{sec:Menge}
The Menge framework originated at the University of North Carolina. Like for \vadere{}, the goal is to facilitate model comparison. For this, the Menge developers provide a very generic framework and invite researchers to contribute to the project.
Menge breaks the simulation down into six sub-problems: \textbf{(1)} Target selection. \textbf{(2)} Plan computation: find the destination by using graphs or potential fields. \textbf{(3)} Plan adaption: use local navigation to find the preferred velocity \textbf{(4)}. Motion synthesis: this means the physical motion of an agent including head, shoulder and feet movement which is not yet addressed within the Menge framework. \textbf{(5)} Environmental queries: identify influencing factors which are in line-of-sight of agent. \textbf{(6)} Crowd systems: simulations of aggregated individuals. 
Compared to \vadere{}, Menge offers but also insists on a software structure which realizes all three levels of pedestrian behavior defined in \cite{hoogendoorn-2004}: the operational (locomotion) layer, the tactical layer, and the strategic layer. This predefined structure is valuable if the model can be mapped onto it but hampering if not. Overhead and additional complexity result in longer development times before a researcher can compare locomotion models.

\paragraph{MomenTUMv2}
\label{sec:MomenTUMv2}
MomenTUMv2 has been developed at Technical University Munich. The focus lies on analyzing and comparing pedestrian behavior models.
Like Menge, the MomenTUMv2 framework implements all levels of pedestrian behavior defined in \cite{hoogendoorn-2004}, that is, the simulation as well as the software itself breaks down into strategic, tactical and operational layers. The strategic layer is responsible for the destination choice of agents. The tactical layer contains four items: \textbf{(1)} navigating to a destination \textbf{(2)} participating (\eg{} in front of a stage) \textbf{(3)} queuing \textbf{(4)} searching unknown locations. The operational layer provides models for walking and waiting agents. Both models can either use a cellular automaton or a force-based model for locomotion. Compared to \vadere{}, and similar to Menge, the three-layered structure in the software introduces development overhead before two locomotion models can be compared.

\paragraph{SUMO}
\label{sec:SUMO}
SUMO is spearheaded by the Institute of Transportation Systems of the German Aerospace Center (DLR). The SUMO simulator allows to evaluate infrastructure changes before implementing them in a real environment. Its scope and its user community are much larger than that of \vadere{} and the three other pedestrian dynamics simulators. We mention it, because in the long run, an interface between SUMO and well-established locomotion models from the pedestrian community would benefit the scientific community.

\section{\vadere{}: A framework to compare different locomotion models}
\label{sec:ResultAndDiscussion}
The \vadere{} project was started in 2010 \cite[p. 23]{sivers-2016b}. Its main intention was and still is to facilitate development and comparison of locomotion models. Therefore, it was designed as a generic framework, but with an eye on keeping it lightweight, so that new locomotion models can be quickly implemented. 
In \cite{seitz-2016}, the software requirements are summarized as shown in \cref{tbl:SoftwareRequirementsForOpenVadereSeitz}. 
This section first introduces \vadere{} from a user's perspective, and then delves into the software architecture for readers who wish to develop with \vadere{}.
\begin{table}[t]
	\centering
	{\scriptsize
		\begin{tabularx}{\linewidth}{ll}
			\toprule
			Functional: &
			\begin{tabular}[t]{ll}
				- & run simulations of pedestrian crowd behavior \\
				- & specify and store parameters in text files with \\
				& a simple format \\
				- & online processing of the simulation \\
				- & online and post-visualization \\
				- & integrated graphical user interface \\
			\end{tabular}\\[3em]
			Non-functional : &
				\begin{tabular}[t]{ll}
				- & only use open-source software \\
				- & run on modern desktop hardware\\
				- & platform independence \\
				- & object-oriented, high-level programming  \\
				& language\\
				- & implement new models without changing the\\
				& framework \\
				- & framework must not impose any model concept or \\
				& structure \\
				- & modular design and architecture \\
				- & re-usability of  basic algorithms and data \\
				& structures \\
			\end{tabular} \\
			\bottomrule
	\end{tabularx}
	 \caption{Software requirements for \vadere{} (taken from \cite[p. 17]{seitz-2016})}
	\label{tbl:SoftwareRequirementsForOpenVadereSeitz}
	}
\end{table}
\subsection{\vadere{} for users}
\label{sec:VadereUsers}

The following section provides an overview of \vadere{} and its optional graphical user interface. The goal is to ease researchers from different domains into the first simulation steps. 
\subsubsection{\vadere{}: overview}
\label{sec:VadereOverview}

\vadere{} is implemented in Java programming language. Thus, it is available for GNU\-/Linux, MacOS and Microsoft Windows. \vadere{} follows the KISS principle \cite{hanik-2006}: keep it simple, stupid. For this reason, \vadere{} reads in simulation parameters, like the topography or an agent's radius, from a human-readable JSON-based text file instead of using a binary format. Also, the simulation results --- usually x and y coordinates for each pedestrian and a time step --- are written to text files. In this way, users can use text editors to create input files for \vadere{} and they can open result files from \vadere{} with 3rd-party software like MATLAB. Furthermore, text files allow users to modify parameters quickly in an automated way. This is essential for studies where parameters must be varied and, thus, 
thousands of simulations must be run. 

\begin{figure}[!h]
    \centering
    \includegraphics[width=\linewidth]{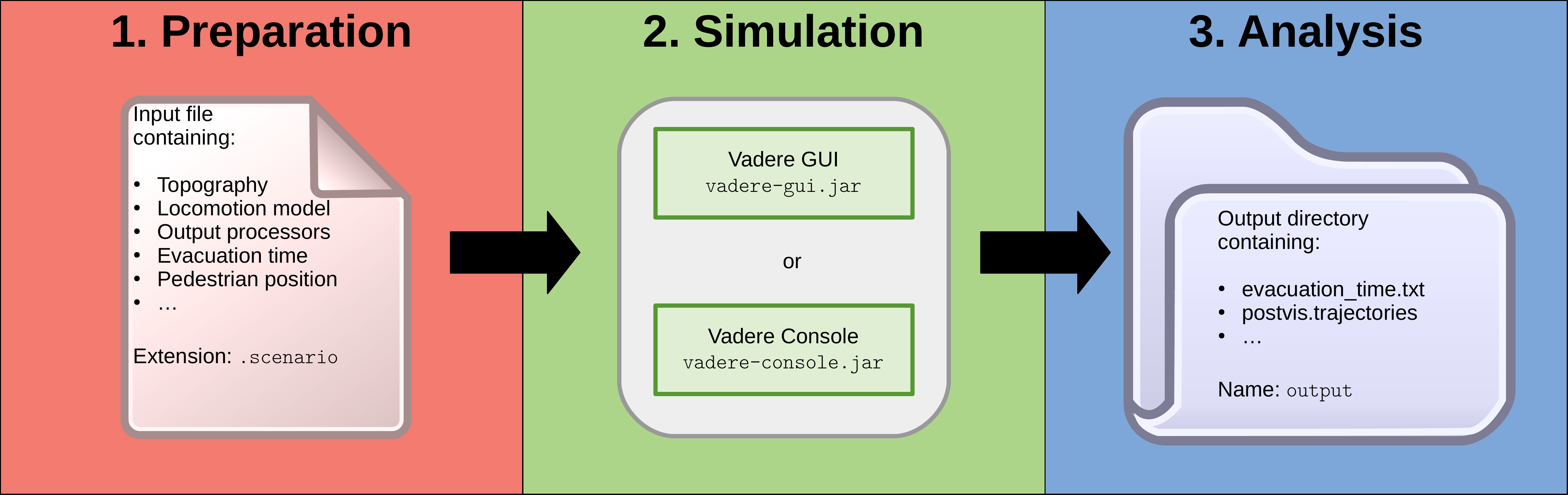}
    \caption{Three steps of a simulation in \vadere{}: (1) Create an input file with extension \code{.scenario} (2) Run \vadere{} GUI or Console with the input file (3) Analyze the output files using \vadere{} GUI or a 3rd-party software.}
    \label{fig:VADEREPreparationSimulationAnalysis}
\end{figure}

Performing simulations with \vadere{} requires three steps visualized in \cref{fig:VADEREPreparationSimulationAnalysis} and summarized below:
\begin{enumerate}
    \item Create an input file for the simulation.
    \item Run the simulation with the input file from (1).
    \item Analyze the simulation results.
\end{enumerate}
The graphical user interface supports all three steps.
Step (1) and (3) can also be carried out with 3rd-party applications that a researcher might be more familiar with.  

When creating input files for the simulation, so-called {\em output processors} can be used to specify what output files should be written. Different output processors exist, \eg{} to write the trajectory of each agent or to write the density at each time step. It is also possible to let multiple output processors write into the same file. Listing \ref{lst:VadereOutputFile} visualizes the structure of an example output file written by \vadere{}.

\begin{lstlisting}[frame=single,caption={Shortened example of an output file written by \vadere{} describing trajectories of three agents.},label={lst:VadereOutputFile}]
timeStep pedestrianId x y
1 1 1.6654341119190224 4.871023637696483
1 2 1.5881140640080735 3.0259345093049403
1 3 1.3134937811396925 2.512808181284221
2 1 2.39821467392553 5.137733950515711
2 2 2.291974452089554 2.7697502789642234
2 3 1.764197128690711 1.7321670841844834
...
\end{lstlisting} 

We think that a highlight for the scientific community lies in our simple, but effective approach to make simulation results completely reproducible. When compiling \vadere{} from sources --- which are under version control using Git --- \vadere{} adds the commit hash to the simulation output files. The commit hash allows to exactly track with which \vadere{} version a simulation was executed.

\subsubsection{\vadere{}: graphical user interface}
\label{sec:VadereGraphicalUserInterface}

\begin{figure}[h]
	\centering
	\includegraphics[width=\linewidth]{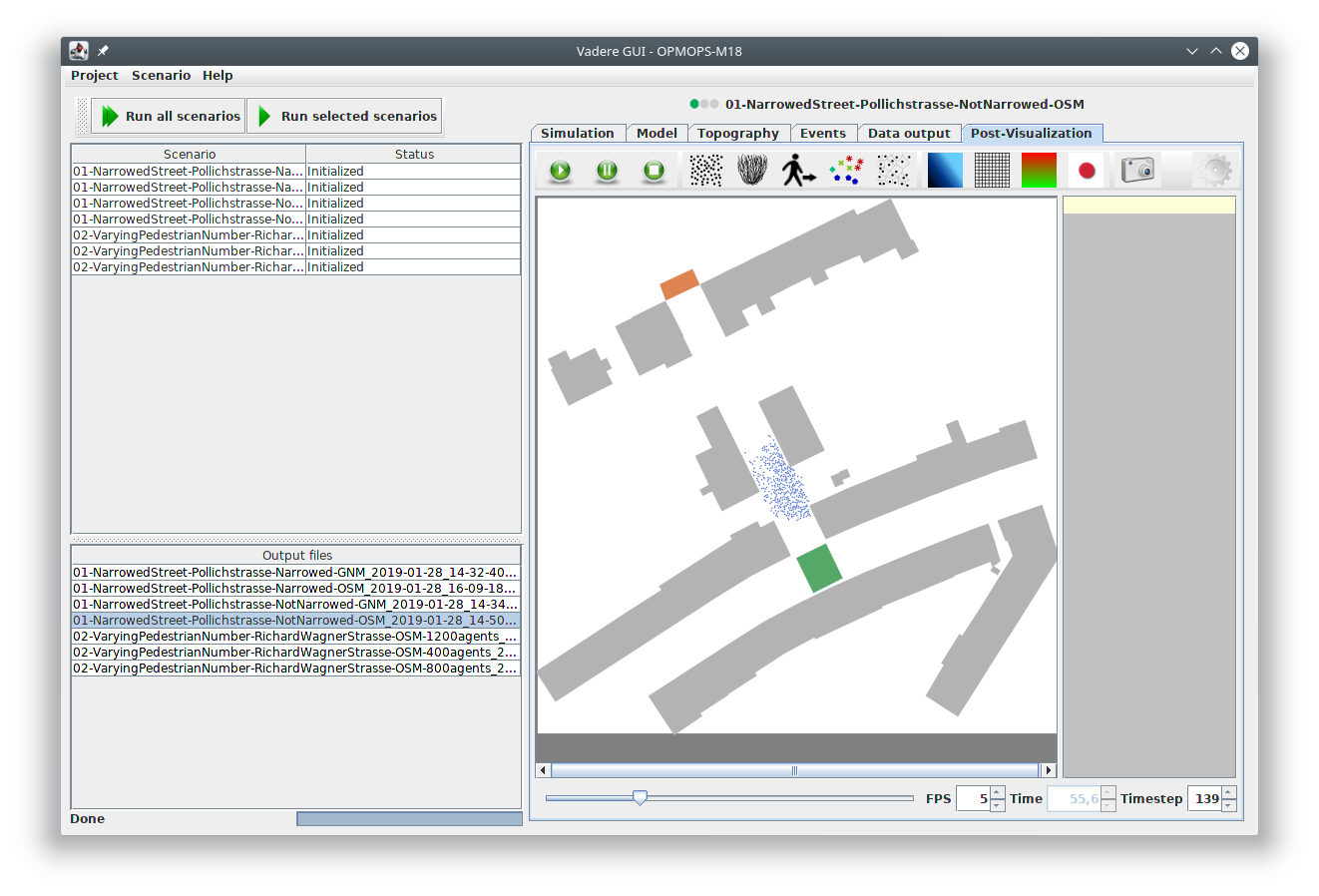}
	\caption{\vadere{} GUI: the top left side lists the available input files, the bottom left side lists the output files and the middle pane contains the online visualization of the simulation.}
	\label{fig:GUIOnlineVisualization}
\end{figure}

\vadere{} includes an optional graphical user interface (GUI) to simplify usage. The GUI has multiple features. Firstly, it provides an overview of the scenario files and the corresponding simulation outputs. Secondly, it offers a simple but effective drawing program to define the topography and to manipulate the attributes of topography elements like agents, sources, targets, obstacles and stairs. Thirdly, during the design of a simulation study, possible problems are identified and transmitted to the user, \eg{} if a pedestrian is defined without a target. The fourth useful GUI feature is the possibility to visualize the simulation run online and offline. Through this monitoring option developers get an immediate visual feedback whether or not agents move \enquote{properly} \cite{gipps-1986}. \cref{fig:GUIOnlineVisualization} shows a screen shot of the \vadere{} GUI. 

We provide an executable file, called \code{vadere-gui.jar} that includes all functionalities. In addition, we provide a console version  called \code{vadere-console.jar}. that should be used to automate simulation runs, for example, when running multiple instances or running \vadere{} on a remote computer.

\subsubsection{\vadere{}: locomotion models supported in \vadere{}}

Currently, \vadere{} supports seven locomotion models, namely the behavioral heuristics model, a simple bio-mechanics model, the gradient navigation model, the optimal steps model, an optimal velocity model, an implementation of Reynolds' steering model and an implementation of the original social force model. The locomotion that emerges from cellular automata is be mimicked by using the optimal steps model (compare \cref{fig:OSMToCA}). This approach is much less efficient than an implementation that makes use of the grid structure to build a true automaton. Therefore, \vadere{} can be used to see whether a cellular automaton model is sufficiently accurate to answer a research question, but it does not do justice to the computational speed that a well-designed cellular automaton achieves.

\subsection{\vadere{} for developers}
\label{sec:VadereDevelopers}
The following section shows \vadere{}'s software architecture in more detail. Moreover, this section sheds light on how to write a generic simulation framework that supports different locomotion models and demands minimal effort for add-ons.  
\begin{figure}[b]
	\centering
	\includegraphics[width=0.5\linewidth]{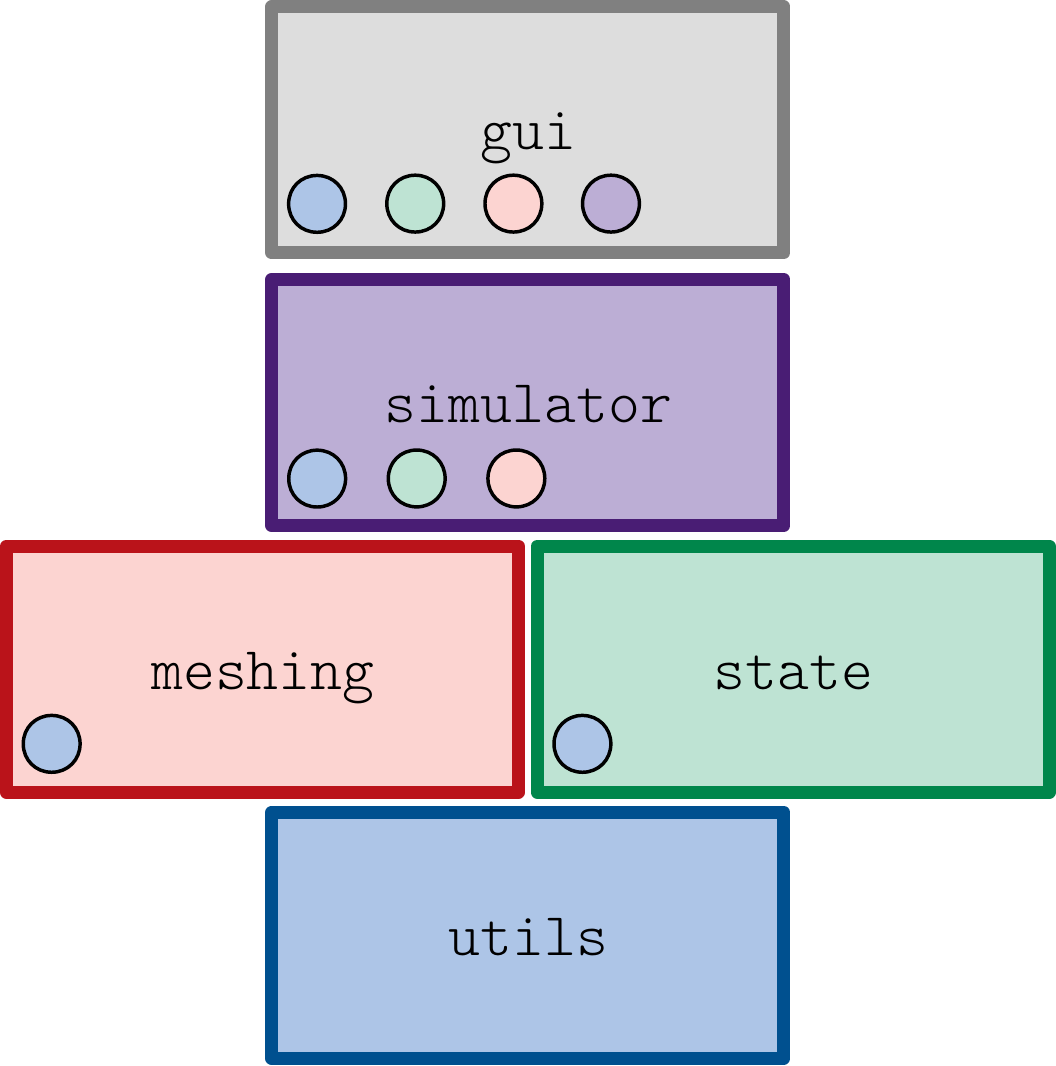}
	\caption{Illustration of \vadere{}'s layered software architecture: A rectangle represents an isolated software module. The software modules in the upper layers depend on modules in the lower layers.}
	\label{fig:VADEREModuleDependenciesLayered}
\end{figure}
\subsubsection{\vadere{}: architecture}
\label{sec:VadereArchitecture}
%
%
\vadere{}'s architecture applies the model view controller (MVC) software pattern \cite{gamma-1994}. Therefore, \vadere{} is divided into three interconnected modules: \code{state}, \code{gui} and \code{simulator}. Moreover, \vadere{} is complemented by two supporting modules: \code{utils} and \code{meshing}. In sum, it is composed of five separated modules depicted in \cref{fig:VADEREModuleDependenciesLayered}.
\begin{figure}[b]
	\centering
	\subfigure[$t = 2,4s$]{\includegraphics[width=0.32\linewidth]{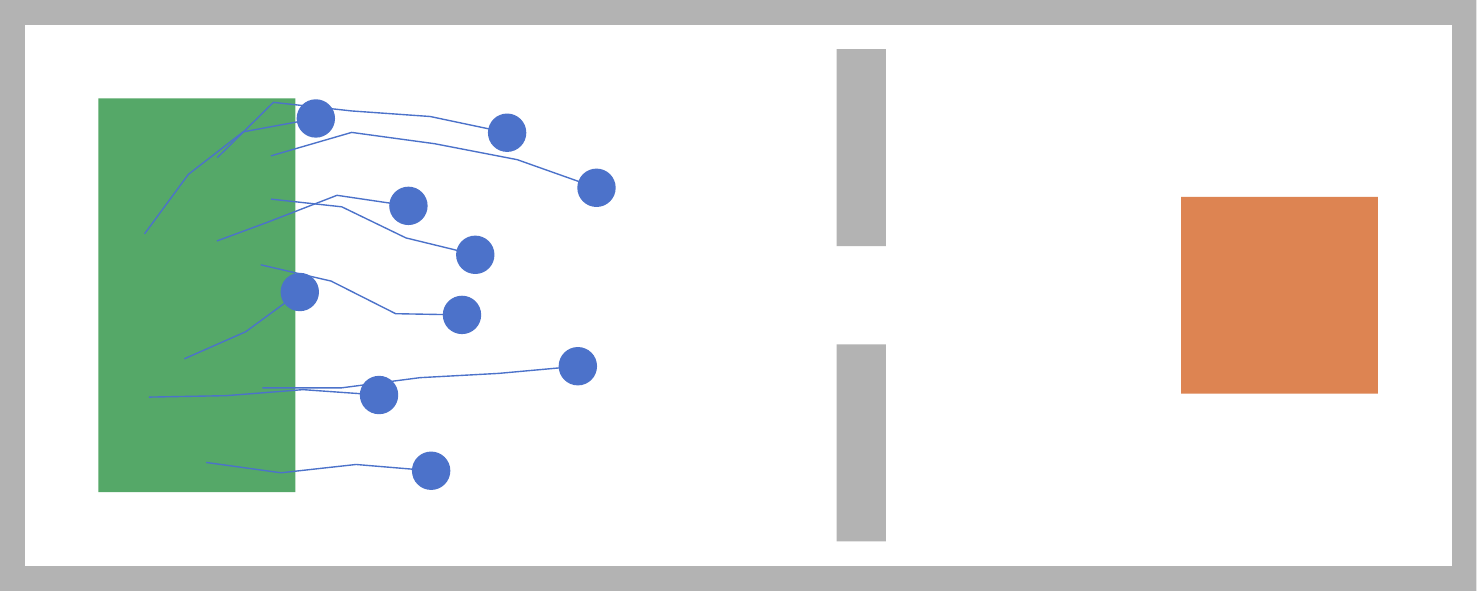}}
	\subfigure[$t =  6s$]{\includegraphics[width=0.32\linewidth]{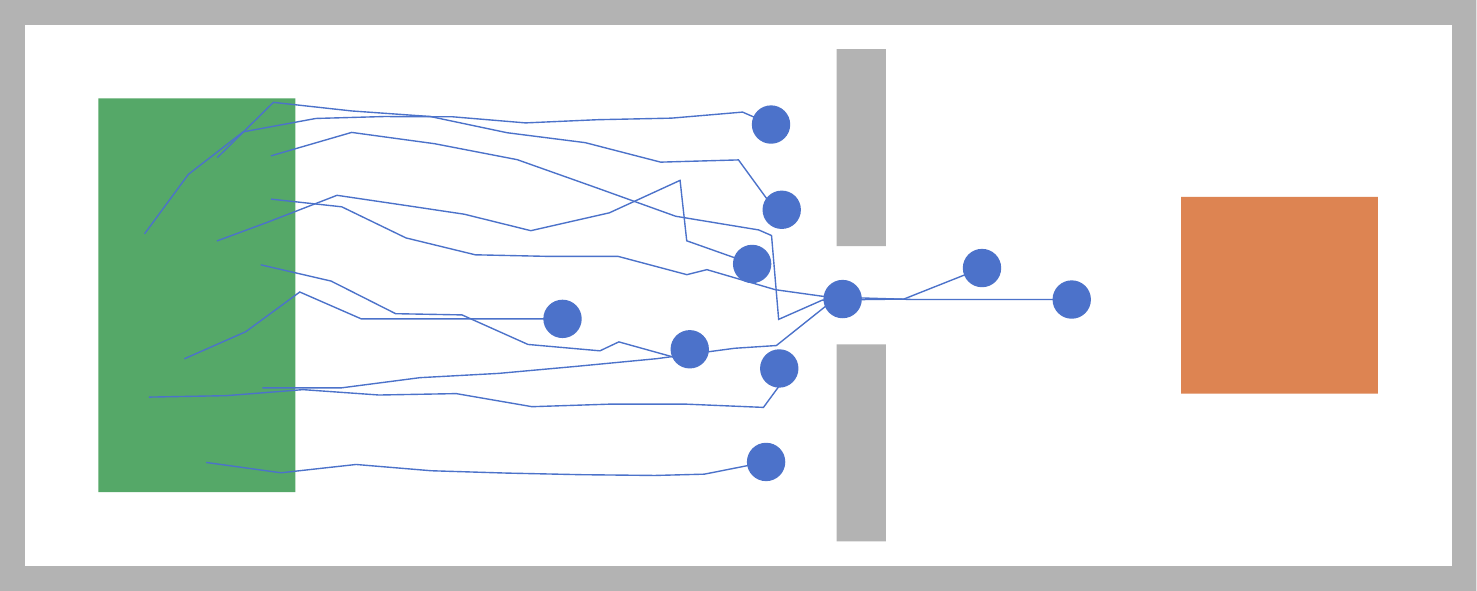}}
	\subfigure[$t = 10,8s$ ]{\includegraphics[width=0.32\linewidth]{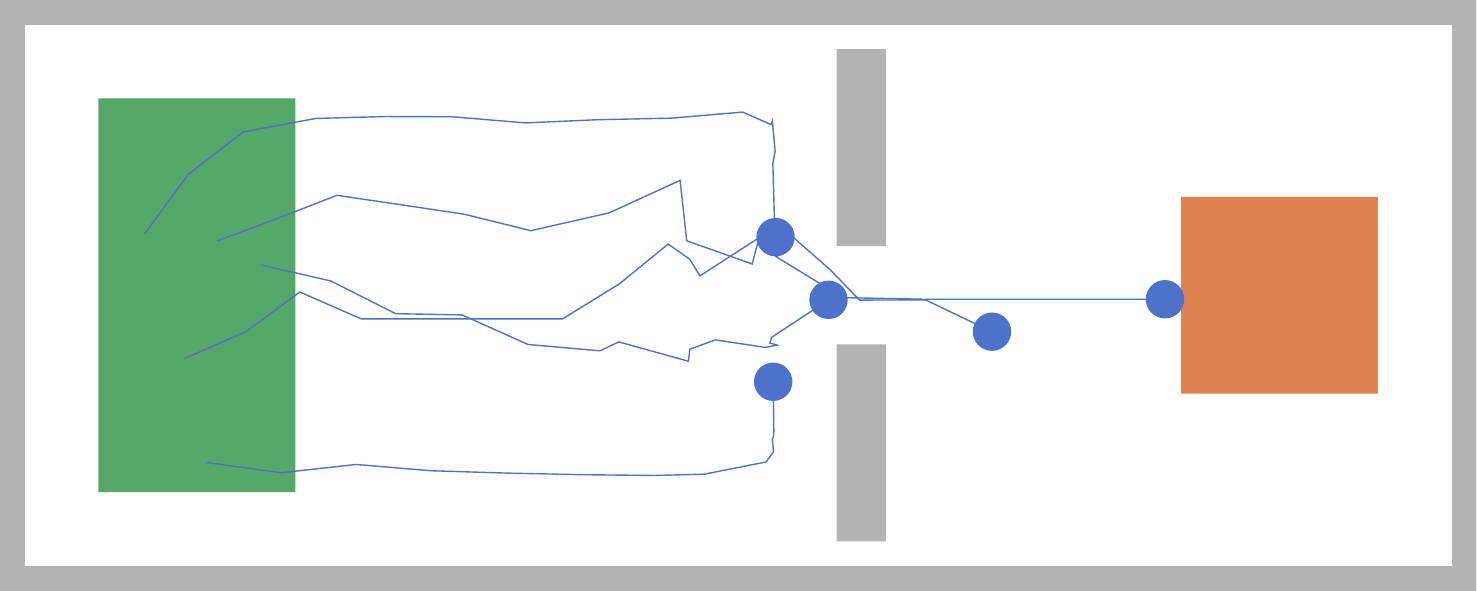}}
	\caption{Agent positions at three different times using the OSM locomotion model.}
	\label{fig:OSMSimulationAtThreeTimeSteps}
\end{figure}

\code{utils} consists of utility classes implementing algorithms required by all other modules such as basic math functions, algorithms for computational geometry, or simple I/O-handling. \code{meshing} is a specialized computational geometry library  that offers methods to mesh a geometry.  The library's mesh algorithms, notably our own EikMesh, are tailored for floor field computation. The library can be used in projects in and outside of the pedestrian dynamics' community. For more information we refer to \cite{zoennchen-2018b}. \code{state} contains topography elements like agents, targets, sources and all attributes, that is, all objects that define the simulation state. \code{simulator} is the most important module
for developing new locomotion models. It contains all locomotion models, the simulation loop itself and all other controller components. The \code{gui} module contains all classes that are part of the graphical user interface. It is optional but recommended, especially for starters, since it improves  usability.
The MVC pattern leads to a clear separation of responsibilities for the three MVC modules within \vadere{}:
\begin{itemize}
    \item Model (\code{state}): the model layer does not contain any logic. Instead, it is the simulation state, \ie{} the composition of agents, sources, obstacles, targets and their corresponding attributes like the number of agents a source creates.
    \item Controller (\code{simulator}): the control layer contains the logic to change objects of the model layer. For instance, to update the x and y coordinates of all agents in each time step. Mainly, the control layer holds implementations for different locomotion models.
    \item View (\code{gui}): The view visualizes the current state of the model objects in form of a GUI. Note that the GUI itself implements the MVC pattern but on another level.
\end{itemize}
\begin{figure}[h!]
	\centering
	\includegraphics[width=\linewidth]{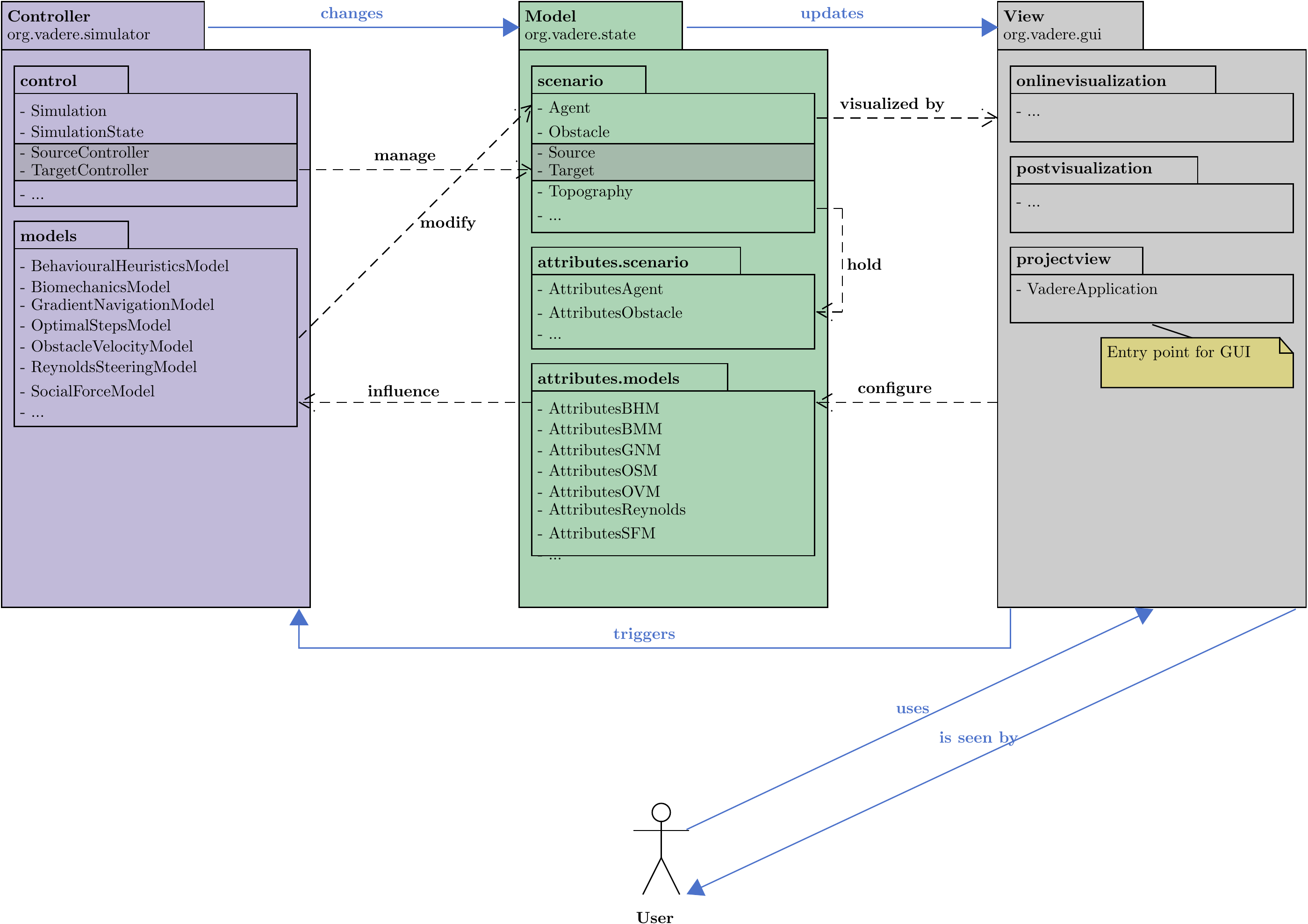}
	\caption{Package diagram showing important classes in \vadere{} and how they are embedded into the MVC pattern. Blue arrows indicate communication between the MVC components. Black arrows show how classes communicate with each other. The controller classes hold the logic to change the model classes which are visualized by the view classes.}
	\label{fig:MVC-VADERE-ImportantClasses}
\end{figure}
\Cref{fig:MVC-VADERE-ImportantClasses} shows a package diagram containing important classes of \vadere{} and how these classes interact with each other.
\subsubsection{The simulation loop}
\label{sec:TheSimulationLoop}
\vadere{}'s core is its simulation loop. When starting a simulation run, \vadere{} enters the simulation loop. In each simulation step, the locomotion model changes the agents' positions by the \code{update} call. See listing \ref{lst:VadereSimulationLoop}. Afterwards, the simulated time is incremented.
Despite \vadere{}'s versatility its core is a single \code{while}-loop. \cref{fig:OSMSimulationAtThreeTimeSteps} shows how the simulation loop updates agents' positions at three different time steps using the OSM locomotion model.
Listing \ref{lst:VadereSimulationLoop} alludes to the \code{SourceController} and the \code{TargetController} controller in a comment line (line 6). The \code{SourceController} spawns new agents inside the topography. The \code{TargetController} removes agents from the topography once they have reached their target. The locomotion model acts on the remaining agents after both operations have been completed.
\begin{lstlisting}[frame=single,caption={Simulation loop of \vadere.},label={lst:VadereSimulationLoop},language=Java]
Model model = ... // loaded from input file
double simulationTime = 0;
...
while (simulationIsRunning) {
    // use "SourceController" and "TargetController"
    ...
    // update the locomotion model
    model.update(simulationTime);
    simulationTime++;
    ...
}
\end{lstlisting}
\subsubsection{Including different locomotion models using the strategy pattern}
\label{sec:IncludingDifferentLocomotionModels}
\begin{figure}[t]
    \centering
    \includegraphics[width=\linewidth]{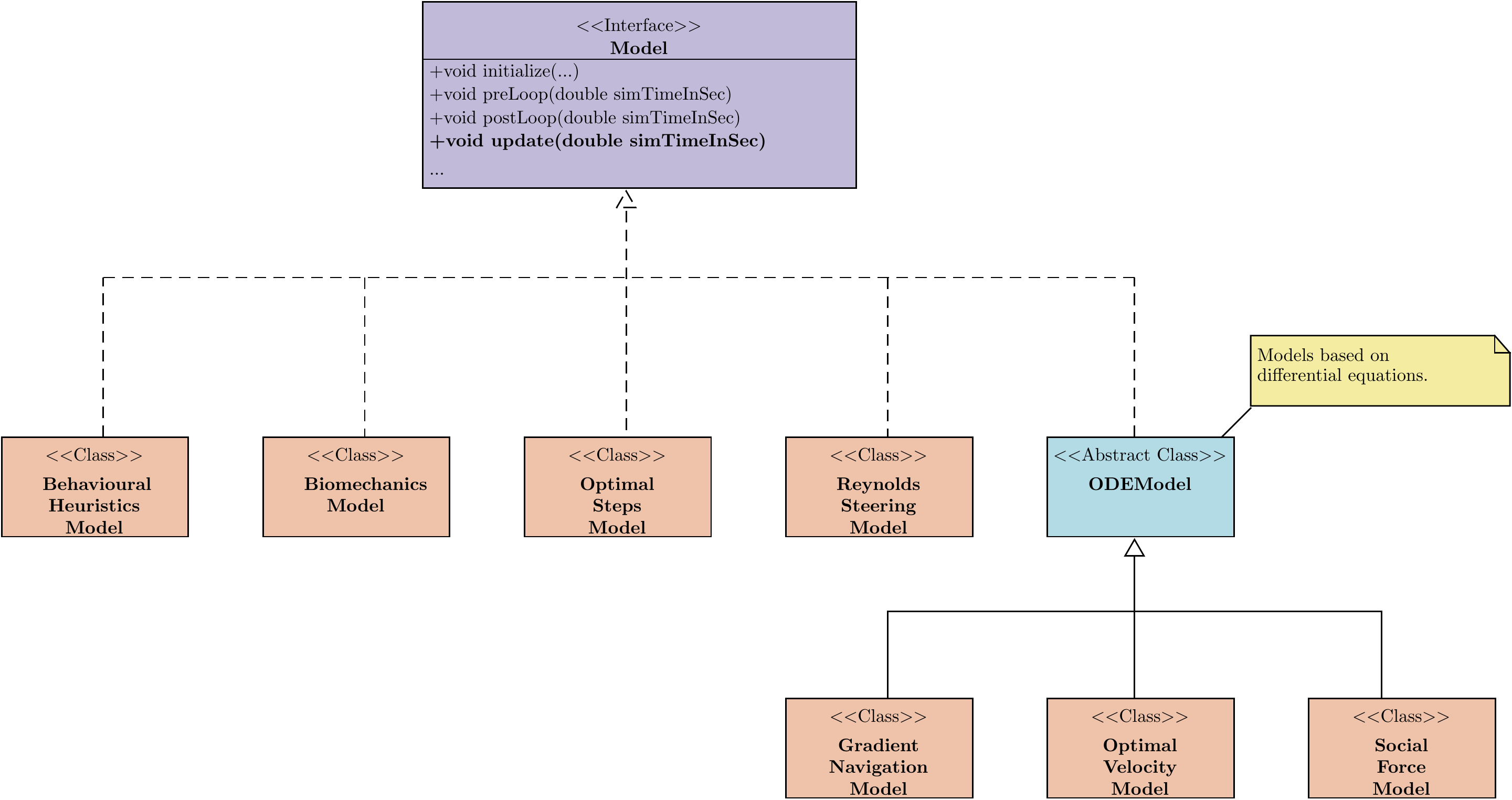}
    \caption{The common locomotion model interface \enquote{Model} and its implementations.}
    \label{fig:VADERE-ClassHierarchy-Models}
\end{figure}

\vadere{}'s goal is to offer a generic framework to support different locomotion models. The challenge is to keep development effort minimal.
The solution idea is to provide an interface. More precisely, 
each locomotion model must implement a common interface that contains four methods. See \cref{fig:VADERE-ClassHierarchy-Models}:
\begin{itemize}
    \item \code{initialize()}: replaces the constructor for the model.
    \item \code{preLoop()}: is called before the simulation loop starts.
    \item \code{postLoop()}: is called after the simulation loop has finished.
    \item \code{update()}: is called during the simulation loop.
\end{itemize}
The simulation loop in listing \ref{lst:VadereSimulationLoop} only holds the generic \code{Model} objects and invokes its \code{update()} method without knowing its concrete implementation. This approach is known as strategy pattern in software engineering. Implementing a new locomotion model in \vadere{} requires just to implement the four methods of the \code{Model} interface. At each time step, the locomotion model has access to the \code{topography} object that encapsulates all required elements like agents, obstacles and targets.

\subsection{Quality assurance: Unit testing and continuous integration}
\label{sec:QualityAssurance}
\begin{figure}[t]
    \centering
    \includegraphics[width=\linewidth]{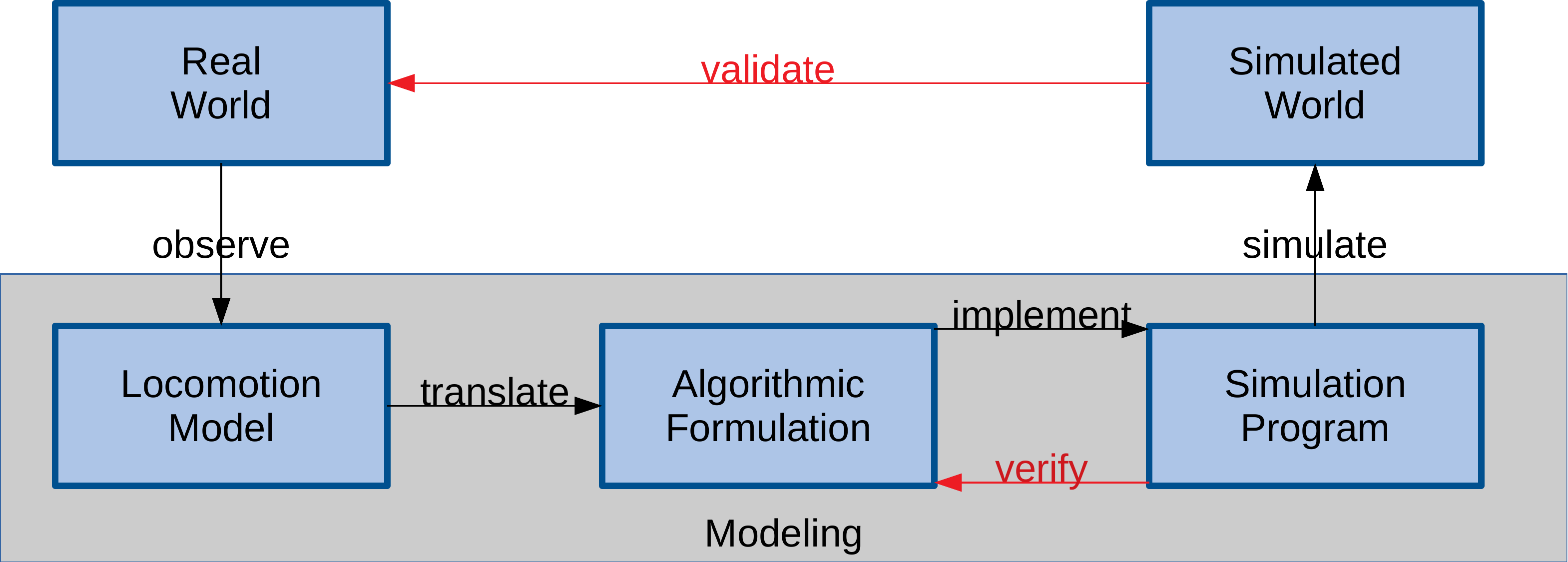}
    \caption{Verification and validation throughout the modeling process: We verify software implementations. We validate the whole model against empirical observations.}
    \label{fig:Modeling-VerificationAndValidation}
\end{figure}

Errors can occur at each stage in the modeling process from 
real-world observations to the mathematical formulation of a locomotion model to its algorithmic formulation, and finally its implementation. See \cref{fig:Modeling-VerificationAndValidation}. Therefore, verification and validation are crucial. 
Through verification we want to assure that the software fulfills the requirements that we formulated, without questioning the requirements.
The latter is the objective of validation where we interpret simulation outcomes as hypotheses derived from our model and compare those to empirical findings. While this can never be more than a failure to falsify one's own model \cite{popper-2002} it is essential for building trust. 
In view of this, we establish a verification and validation process as integral part of \vadere{}.
\begin{lstlisting}[frame=single,caption={The test class \code{TestGeometry} tests the code in class \code{Geometry}.},label={lst:VadereUnitTestExample},language=Java]
public class TestGeometry {
    ...
    @Test
    public void testLineCircleIntersectionZeroResults() {
        VCircle circle = new VCircle(1, 1, 1); // x,y,radius
        VLine line = new VLine(3,0, 4, 1); // x1,y1,x2,y2
        
        // Method under test.
        VPoint[] intersectionPoints = 
        	GeometryUtils.intersection(line, circle);
        assertTrue(intersectionPoints.length == 0);
    }
    ...
}
 
\end{lstlisting}
\subsubsection{Unit Testing}
\label{sec:UnitTesting}
Crucial code is best developed in a test-driven manner to enforce correctness of written source code. Listing \ref{lst:VadereUnitTestExample} shows the principle of unit testing in the \vadere{} project with an example. We use the Java unit testing framework JUnit \cite{junit-2018}. \vadere{}'s test suite currently covers 24\%\footnote{Retrieve the detailed test report at \url{https://gitlab.lrz.de/vadere/vadere/-/jobs/370587}.} of the productive source code excluding the \code{View} components. See \cref{fig:MVC-VADERE-ImportantClasses}. One could argue that this is a low coverage compared to industrial software \cite{iso-2018-26262-6}, but we believe it is sufficient for a research simulator that needs to stay flexible. Also, we focus on testing core classes to make the simulations reliable while keeping development fast.  

\subsubsection{Continuous integration and deployment}
\label{sec:ContinuousIntegrationAndDeployment}

\begin{figure}[!h]
    \centering
    \includegraphics[width=\linewidth]{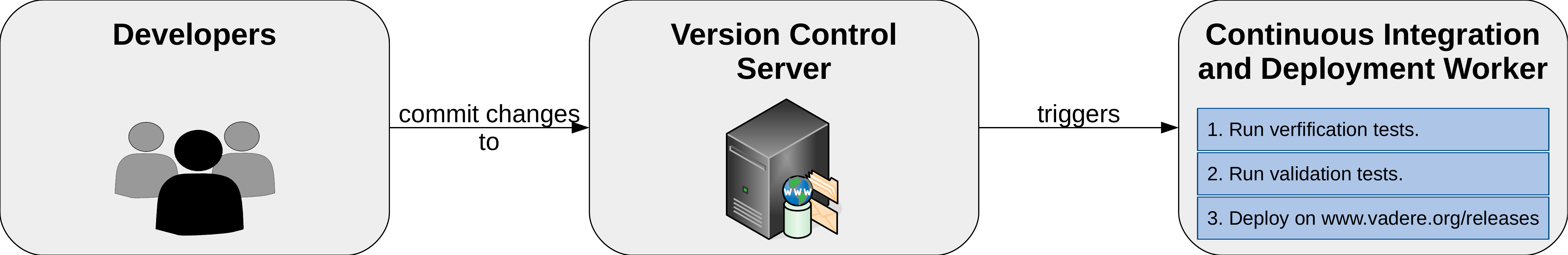}
    \caption{The three actors in the continuous integration and deployment cycle: Developers commit changes to the version control server. The version control server triggers worker computers which validate, verify and deploy \vadere{}.}
    \label{fig:VerificationAndValidation-Steps}
\end{figure}

Our second strategy for error reduction is so-called continuous integration.
In our case this means that a test pipeline is invoked each time the \vadere{} source code is modified and pushed to the source code repository. First, the test pipeline performs all unit tests. This represents the code verification. Then, special test scenarios which are defined by the RiMEA ``Guideline for Microscopic Evacuation Analysis" 
\cite[pp. 38--51]{rimea-2016} are executed. The guideline defines 15 test cases for evacuation scenarios which shall be passed by a pedestrian dynamics' simulator. Additionally, we define 16 scenarios based on experiments published and evaluated in \cite{zhang-2011} and compare simulation results to real world data. 
Test scenarios serve partly as model verification and partly as model validation. Currently, for the OSM locomotion, at least 49 test runs with different parameter sets are executed. 
Finally, the verified and validated executable of the \vadere{} simulator is placed on the website \url{http://www.vadere.org/releases/}. The website hosts \vadere{} versions for GNU/Linux and Microsoft Windows. We use the web-based Git repository manager GitLab \cite{gitlabcontributors-2018} to manage our code base. With GitLab's pipeline feature and self-written Python scripts --- which are kept in the repository as well --- we are able to automate the three steps: verification, validation and deployment of \vadere{}. \cref{fig:VerificationAndValidation-Steps} summarizes all steps that are carried out during the continuous integration and deployment cycle. GitLab reports any errors during the two pipeline stages via email to the developers and provides error details through the web interface. Furthermore, we use GitLab's issue tracker to create feature requests and bug reports. This makes the development process transparent.

\section{Conclusion}
\label{sec:Conclusion}

In this paper, we presented the free and open-source simulation framework \vadere{} for pedestrian dynamics. \vadere{} contains implementations of several locomotion models, among them the cellular automata, the social force model and the optimal steps model to facilitate model comparison. We described \vadere{}'s architecture and showed how to integrate new locomotion models in a straightforward way. Each model's implementation is verified based on JUnit tests and validated against the standard test scenarios from the RiMEA guidelines. In addition, for any change in the code both verification and validation tests are automatically run and results are communicated to the developers. This mechanism assures a high source code quality and gives developers an immediate feedback when a result is considered implausible by the scientific community.

Our vision of \vadere{} is a tool that encourages research within and is used by the whole interdisciplinary pedestrian dynamics community instead of just serving the purposes of our research group. We encourage each member of this interdisciplinary community to use \vadere{} in the way that suits him or her: as a user who  conducts and analyzes simulation experiments, as a modeler who introduces new models or as a software developer who improves and extends the software. 

\section{Acknowledgments}
We thank the research office (FORWIN) of the Munich University of Applied Sciences and the Faculty Graduate Center CeDoSIA of TUM Graduate School at Technical University of Munich for their support.

\paragraph{Authors' contributions} B.\,K. drafted the article. B.\,Z. revised the article, especially \cref{sec:intro} and \ref{sec:LocomotionModels}. M.\,G. critically reviewed the article and provided input for the conclusion. G.\,K. critically revised the article and added the abstract and gave final approval for publication.

\paragraph{Funding}
B.\,K. is supported by the German Federal Ministry of Education and Research through the project OPMoPS to study organized pedestrian movement in public spaces (grant no. 13N14562). B.\,Z. and M.\,G. are supported by the German Federal Ministry of Education and Research through the project S2UCRE to study the acceleration of microscopic pedestrian simulations by designing efficient and parallel algorithms (grant no. 13N14463).

\paragraph{\vadere{} contributors (in alphabetical order)}
Core developers: Felix Dietrich, Michael Seitz, Isabella von Sivers, Benedikt Z\"{o}nnchen; Contributors: Florian Albrecht, Benjamin Degenhart, Marion G\"{o}del, Benedikt Kleinmeier, Daniel Lehmberg, Jakob Sch\"{o}ttl, Stefan Schuhb\"{a}ck, Swen Stemmer, Mario Teixeira Parente, Peter Zarnitz.

\bibliographystyle{plain} 
\bibliography{Literature} 


\newpage
\appendix
\section{Appendix}
\label{app:1}

\subsection{Obtain lines of code (LOCs) for different simulators}
\label{app:ObtainLOCsForDifferentSimulators}

The lines of code exclude unit tests, blank lines and comments. The "cloc" software tool
\footnote{\url{https://github.com/AlDanial/cloc}}
 version 1.74 was used to obtain the lines of code. The hash contained in the \code{--report-file} indicates the analyzed simulator version according to the Git version control system.

\begin{lstlisting}
JuPedSim:
  cloc
  --exclude-dir=Utest
  --exclude-lang=XML
  --report-file=jupedsim-d942c947-cloc_report.txt
  jpscore/ jpseditor/ jpsreport/ jpsvis/
    .
Menge:
  cloc
  --match-d=src
  --exclude_dir=test
  --report-file=menge-menge-c3eb429-cloc_report.txt
  .
MomenTUMv2:
  cloc
  --exclude-dir=momentum-documentation,tests
  --exclude-lang=HTML,CSS,XML
  --report-file=momentumv2-55c8f3a-cloc_report.txt
  .
SUMO:
  cloc
  --match-d=src
  --report-file=sumo-1.0.1-cloc_report.txt
  .
Vadere:
  cloc
  --exclude-dir=tests
  --exclude-lang=JSON --report-file=vadere-87b4fe32-cloc_report.txt
  .
\end{lstlisting}

\end{document}